\documentclass[10pt,twocolumn,letterpaper]{article}
\usepackage[dvipsnames,table]{xcolor}
\usepackage[pagenumbers]{cvpr} 

\usepackage{acro}
\usepackage{multirow}
\usepackage{array}
\usepackage{float}

\definecolor{id}{RGB}{173, 216, 230}
\definecolor{near}{RGB}{144, 238, 144}
\definecolor{far}{RGB}{255, 255, 224}
\definecolor{csid}{RGB}{255, 182, 193}
\definecolor{gr}{RGB}{235, 235, 235}
\definecolor{dark_gr}{RGB}{0, 0, 0}

\definecolor{Classification}{RGB}{31, 119, 180}
\definecolor{Feature}{RGB}{255, 127, 14}
\definecolor{Combined}{RGB}{44, 160, 44}

\definecolor{Classification_light}{RGB}{183, 219, 242} 
\definecolor{Feature_light}{RGB}{255, 210, 170} 
\definecolor{Combined_light}{RGB}{188, 237, 188} 
\definecolor{Gray_light}{RGB}{222, 222, 222}

\usepackage{siunitx}
\DeclareSIUnit{\px}{px}
\DeclareSIUnit{\vo}{vo}

\usepackage[]{tikz}							
\usetikzlibrary{shapes.arrows}
\usetikzlibrary{positioning}
\usetikzlibrary{tikzmark} 
\usetikzlibrary{calc}
\usetikzlibrary{patterns}     

\usepackage{float} 

\usepackage{tabularx}
\definecolor{cvprblue}{rgb}{0.21,0.49,0.74}
\usepackage[pagebackref,breaklinks,colorlinks,citecolor=cvprblue]{hyperref}
\usepackage{newtxtext,newtxmath}

\def\paperID{10513} 
\def\confName{CVPR}
\def\confYear{2025}

\title{
\begin{center}
\vspace*{-2.0cm} 
\fbox{\parbox{0.93\textwidth}{
\small \textbf{Update (October 2025):} 
This version includes a \hyperref[sec:corrigendum]{Corrigendum} after the Appendix, correcting reported results for ViM and NNGuide. 
Please refer to the corrigendum section for details and updated tables.}}
\end{center}
\vspace{1em}
OpenMIBOOD: Open Medical Imaging Benchmarks for\\Out-Of-Distribution Detection}

\author{Max Gutbrod$^{1,2}$, David Rauber$^1$, Danilo Weber Nunes$^{1,2}$, Christoph Palm$^{1,2}$ \\
{\small $^1$Regensburg Medical Image Computing (ReMIC), OTH Regensburg, Regensburg, 93053, Germany}\\
{\small $^2$Regensburg Center of Health Sciences and Technology (RCHST), OTH Regensburg, Regensburg, 93053, Germany}\\
{\tt\small \{max.gutbrod,christoph.palm\}@oth-regensburg.de}
}

\DeclareAcronym{AnD}{
    short = AnD,
    long = anomaly detection
}

\DeclareAcronym{AD}{
    short = AD,
    long = Alzheimer's disease
}

\DeclareAcronym{ADNI}{
    short = ADNI,
    long = Alzheimer's Disease Neuroimaging Initiative
}

\DeclareAcronym{AI}{
	short = AI,
	long = Artificial Intelligence
}

\DeclareAcronym{ATLAS}{
    short = ATLAS,
    long = ATLAS
}

\DeclareAcronym{AUROC}{
    short = AUROC,
    long = area under the receiver operating characteristic curve
}

\DeclareAcronym{AUPR}{
    short = AUPR,
    long = area under the Precision-Recall curve
}

\DeclareAcronym{AUPRIN}{
    short = $\text{AUPR}_\text{IN}$,
    long = $\text{AUPR}_\text{IN}$
}

\DeclareAcronym{AUPROUT}{
    short = $\text{AUPR}_\text{OUT}$,
    long = $\text{AUPR}_\text{OUT}$
}

\DeclareAcronym{BRATS}{
    short = BraTS,
    long = BraTS Challenge
}

\DeclareAcronym{CATARACTS}{
    short = CATARACTS,
    long = CATARACTS
}

\DeclareAcronym{CHAOS}{
    short = CHAOS,
    long = CHAOS Challenge
}

\DeclareAcronym{cholec}{
    short = Cholec80,
    long = Cholec80
}

\DeclareAcronym{csID}{
	short = cs-ID,
	long = covariate-shifted in-distribution
}

\DeclareAcronym{CN}{
    short = CN,
    long = cognitively normal
}

\DeclareAcronym{CE}{
    short = CE,
    long = cross-entropy
}

\DeclareAcronym{EndoSeg15}{
    short = EndoSeg15,
    long = instrument segmentation and tracking dataset
}

\DeclareAcronym{EndoSeg18}{
    short = EndoSeg18,
    long = robotic scene segmentation dataset
}

\DeclareAcronym{CT}{
    short = CT,
    long = Computed Tomography
}

\DeclareAcronym{EndoVis}{
    short = EndoVis,
    long = Endoscopic Vision Challenge
}

\DeclareAcronym{EndoVis2015}{
    short = EndoVis 2015,
    long = Endoscopic Vision Challenge 2015
}

\DeclareAcronym{EndoVis2018}{
    short = EndoVis 2018,
    long = Endoscopic Vision Challenge 2018
}

\DeclareAcronym{FNAC}{
    short = FNAC,
    long = fine needle aspiration cytology
}

\DeclareAcronym{HE}{
    short = H\,\&\,E,
    long = Hematoxylin \& Eosin
}

\DeclareAcronym{imposter}{
    short = imposter,
    long = mitotic cell lookalike
}

\DeclareAcronym{FPR}{
    short = FPR,
    long = false positive rate
}

\DeclareAcronym{TPR}{
    short = TPR,
    long = true positive rate
}

\DeclareAcronym{MSD-H}{
    short = MSD-H,
    long = Medical Segmentation Decathlon Heart
}

\DeclareAcronym{ID}{
    short = ID,
    long = in-distribution
}

\DeclareAcronym{IN1k}{
    short = ImageNet1k, 
    long = ImageNet1k
}

\DeclareAcronym{KVASIR}{
    short = Kvasir-SEG,
    long = Kvasir-Seg
}

\DeclareAcronym{ML}{
	short = ML,
	long = Machine Learning
}

\DeclareAcronym{MRI}{
    short = MRI,
    short-plural-form = MRIs,
    long = Magnetic Resonance Imaging
}

\DeclareAcronym{NN}{
	short = NN,
	long = Neural Network
}

\DeclareAcronym{OTHR}{
	short = OTH,
	long = Ostbayerische Technische Hochschule Regensburg
}

\DeclareAcronym{OOD}{
    short = OOD,
    long = out-of-distribution
}

\DeclareAcronym{OODD}{
    short = OODD,
    long = out-of-distribution detection
}

\DeclareAcronym{MCI}{
    short = MCI,
    long = mild cognitive impairment
}

\DeclareAcronym{MIB}{
    short = MIB,
    long = Medical Imaging Benchmark
}

\DeclareAcronym{MIDOG}{
    short = MIDOG,
    long = MIDOG Challenge
}

\DeclareAcronym{MS}{
    short = MS,
    long = Multiple sclerosis lesions
}

\DeclareAcronym{nOOD}{
    short = near-OOD,
    long = near-OOD
}

\DeclareAcronym{fOOD}{
    short = far-OOD,
    long = far-OOD
}

\DeclareAcronym{OASIS3}{
    short = OASIS-3,
    long = Open Access Series of Imaging Studies 3 dataset
}

\DeclareAcronym{OBOOM}{
    short = OpenMIBOOD,
    long = Open Medical Imaging Benchmarks for Out-Of-Distribution Detection
}

\DeclareAcronym{OSR}{
    short = OSR,
    long = open set recognition
}

\DeclareAcronym{PHAKIR}{
    short = PhaKIR,
    long = PhaKIR
}

\DeclareAcronym{SGD}{
    short = SGD,
    long = stochastic gradient descent
}

\DeclareAcronym{T1w}{
    short = T1w,
    long = T1-weighted
}

\DeclareAcronym{T2w}{
    short = T2w,
    long = T2-weighted
}

\DeclareAcronym{T2w-lr}{
    short = T2w-lr,
    long = low-res T2w
}

\DeclareAcronym{UQ}{
    short = UQ,
    long = uncertainty quantification
}

\begin{document}

\maketitle

\input{figures/datasets}

\begin{abstract}
The growing reliance on \ac{AI} in critical domains such as healthcare demands robust mechanisms to ensure the trustworthiness of these systems, especially when faced with unexpected or anomalous inputs. 
This paper introduces the \ac{OBOOM}, a comprehensive framework for evaluating \ac{OOD} detection methods specifically in medical imaging contexts. 
\ac{OBOOM} includes three benchmarks from diverse medical domains, encompassing 14 datasets divided into \acl{csID}, \acl{nOOD}, and \acl{fOOD} categories.
We evaluate 24 post-hoc methods across these benchmarks, providing a standardized reference to advance the development and fair comparison of \ac{OOD} detection methods.
Results reveal that findings from broad-scale \ac{OOD} benchmarks in natural image domains do not translate to medical applications, underscoring the critical need for such benchmarks in the medical field.
By mitigating the risk of exposing \ac{AI} models to inputs outside their training distribution, \ac{OBOOM} aims to support the advancement of reliable and trustworthy AI systems in healthcare.
The repository is available at \url{https://github.com/remic-othr/OpenMIBOOD}.
\end{abstract}    
\section{Introduction}
\label{sec:intro}
The rapid proliferation of \ac{AI} systems in everyday life raises concerns about their trustworthiness.
According to the technical specification ISO/IEC TS5723:2022~\cite{ISO_TS5723:2022}, an important aspect of trustworthiness concerns a system's ability to handle unexpected inputs.
Such inputs pose significant risks in safety-critical areas, including autonomous driving~\cite{Bogdoll_2022_CVPR} and the healthcare sector~\cite{yang2024generalized}.
Most AI systems are still trained and evaluated under the assumption that the training data distribution -- referred to as \ac{ID} data -- matches the distribution of data encountered post-deployment. 
However, this assumption overlooks the possibility of encountering data from unknown and unseen distributions, known as \acf{OOD} data.
When confronted with such data, \ac{AI} models often exhibit high confidence in their predictions, even when these predictions are entirely incorrect~\cite{goodfellow2015exp}.
Such behavior can result in silent and potentially catastrophic failures, particularly in high-stakes domains like healthcare, where erroneous predictions could directly impact patient safety.
To address this, \ac{OOD} detection methods help distinguish \ac{ID} from \ac{OOD} inputs, allowing models to flag or discard unreliable predictions or refer them for human review.
Since 2016, numerous \ac{OOD} detection methods have emerged~\cite{yang2022openood}, but a unified, comprehensive benchmark was lacking.
Yang \etal addressed this by introducing OpenOOD~\cite{yang2022openood}, an open-source codebase that integrates related fields and provides a consistent evaluation framework.

In their framework, \ac{OOD} is divided into two categories.
The more challenging \ac{nOOD}, characterized by exhibiting similar semantics or styles compared to the \ac{ID} datasets, making it particularly difficult to distinguish them.
And the less challenging \ac{fOOD}, typically featuring different semantic labels and styles~\cite{fang2022is, yang2022openood}.

In a follow-up version, OpenOOD v1.5~\cite{zhang2024openood}, Zhang \etal expanded their framework to encompass large-scale and full-spectrum \ac{OOD} detection~\cite{yang2023full}, covering over 50 methods.
In this full-spectrum setting, the authors merged \ac{ID} with \acf{csID} data -- data, retaining the same labels and class relationships as the \ac{ID} data while exhibiting variations in the distribution of input features.
They argue that this approach evaluates not only \ac{OOD} detection capabilities but also the generalization performance of an \ac{AI} system.
While this approach has undeniable value, we argue for analyzing \ac{OOD} detection and model generalization independently to yield more precise insights.
Furthermore, depending on the application, distinguishing between \ac{ID} and \ac{csID} can be just as critical.
For instance, in medical imaging, where training data is often limited or inaccessible due to privacy constraints, achieving robust generalization may be challenging.
In such cases, focusing on detecting \ac{csID} data can prove more impactful than solely aiming to enhance generalization performance.

We propose that, rather than combining \ac{ID} and \ac{csID} data, \ac{OOD} detection methods should be rigorously evaluated based on their capacity to differentiate between these two categories.
This is particularly important in domains like healthcare, where the ability to identify subtle covariate shifts in data is essential to maintaining reliability.

To address the absence of medical images in the OpenOOD benchmark, we introduce three \acp{MIB}, which consist of \num{14} datasets organized into three distinct \ac{OOD} evaluation settings named after their \ac{ID} datasets: MIDOG~\cite{aubreville2023comprehensive}, \acl{PHAKIR}~\cite{rueckert2024miccai}, and \acs{OASIS3}~\cite{lamontagne2019oasis}. 
An overview of these benchmarks and their involved domain shifts is provided in \cref{fig:datasets}.

\ac{OOD} detection methods are typically divided into two main types: post-hoc methods applied solely during inference, and methods requiring additional training steps. 
This work primarily focuses on post-hoc methods, valued for their ease of integration and model-agnostic nature. 
Besides these advantages, the analysis of Yang \etal showed that post-hoc methods are no worse than methods that require additional training steps~\cite{yang2022openood}.

Following an overview of the current benchmarks for \ac{OOD} detection methods in the medical field (\cref{sec:related}), we describe all evaluation settings, their underlying datasets, and the employed evaluation metrics (\cref{sec:datasets}). 
Subsequently, we present the results of all evaluated methods, together with an in-depth discussion of these results (\cref{sec:experiments}) and a summary of key findings and future research directions (\cref{sec:conclusion}).

Our main contributions are as follows:
\begin{itemize}
    \item We introduce the most comprehensive \ac{OOD} benchmarks for medical imaging across histopathology, endoscopy, and \ac{MRI} domains, evaluating \num{24} post-hoc methods and revealing gaps in current approaches for certain domain shifts in medical data.
    \item We show that selecting top-performing methods from benchmarks based on natural images likely leads to suboptimal performance on medical imaging data.
    \item To reproduce our results, we provide a comprehensive codebase including all evaluated \ac{OOD} detection methods, along with download links for the employed machine learning models and links to all utilized datasets.
\end{itemize}
\section{Related Work}
\label{sec:related}
\ac{OOD} detection is central to medical image processing, as it ensures that deep learning models can appropriately handle and respond to data that falls outside their training distribution. 
Despite its importance, only a limited number of publications have systematically evaluated \ac{OOD} methods in this specific domain \cite{zimmerer2022mood, cao2020benchmark, vasiliuk2023limitations}. Instead, most studies have prioritized the development of new methods~\cite{hong2024out}.

The Medical Out-of-Distribution Analysis Challenge~\cite{zimmerer2022mood} serves as a major benchmark for evaluating \ac{OOD} and anomaly detection in medical images. 
Its \ac{ID} datasets include \ac{MRI} scans of healthy adults and \ac{CT} scans from patients over 50 years old referred for colonoscopy. 
The \ac{OOD} test data feature synthetic image corruptions, alterations, and destructions, aligning with the \ac{csID} category of~\cite{zhang2024openood}, as well as images of unseen medical conditions that correspond to \ac{nOOD}. 
The challenge results reveal a substantial variance in the performance of \ac{OOD} detection methods across different types of anomalies, indicating that none of the current approaches are sufficiently robust for immediate clinical deployment. 
However, this challenge has limitations: it does not comprehensively evaluate \ac{OOD} detection methods across diverse medical imaging domains, nor does it offer a unified framework to guide and facilitate future research and development in the field.

Cao \etal~\cite{cao2020benchmark} categorize \ac{OOD} examples into three use cases: rejecting improperly preprocessed inputs (\ie \ac{csID}), rejecting inputs that contain unseen conditions or artifacts (\ie \ac{nOOD}), and rejecting unrelated inputs (\ie \ac{fOOD}), such as cat pictures. 
Their evaluation spans eight post-hoc \ac{OOD} detection methods and eight methods requiring additional training steps applied across three medical imaging contexts: chest X-ray, fundus imaging, and histology.
However, the scope of their evaluation is notably limited, encompassing only basic post-hoc approaches.
Additionally, the quality of the curated datasets presents challenges. 
For instance, evaluations involving natural images have limited relevance in a medical context.
Furthermore, most evaluation settings rely on a single \ac{OOD} dataset or even omit the essential context of \ac{nOOD} datasets.

In~\cite{vasiliuk2023limitations} the effectiveness of \ac{OOD} detection methods is evaluated with a focus on medical 3D image segmentation. 
The \ac{ID} datasets include \ac{CT} images showing lung nodules and \ac{MRI} images depicting benign brain tumors. 
Various \ac{OOD} scenarios are evaluated, incorporating distribution shifts caused by different imaging devices and synthetic image corruptions (\ie \ac{csID}), unseen diseases (\ie \ac{nOOD}) and entirely different organs (\ie \ac{fOOD}). 
The study evaluates three post-hoc methods and three methods that change the model architecture or training process. 
While the work contributes valuable insights into \ac{OOD} detection within the context of 3D medical image segmentation, its scope remains limited. 
Notably, the study primarily emphasizes the evaluation of the author's data-centric \ac{OOD} detection method.

In addition to studies focusing on the evaluation of various \ac{OOD} detection methods, Hong \etal~\cite{hong2024out} offer a comprehensive review of the \ac{OOD} detection landscape within medical image analysis.
They argue that the differentiation between \ac{csID} and \ac{nOOD} from OpenOOD is not entirely applicable for medical images. 
For example, while a classifier trained on X-ray images might still perform accurately on low-contrast X-rays, its performance is likely to degrade significantly when applied to entirely different modalities, such as \ac{CT} scans~\cite{hong2024out}.
Under the strict taxonomy outlined in \cite{yang2022openood,zhang2024openood}, both shifts would technically fall under the \ac{csID} category, as class labels remain unchanged, despite the significant differences in modality.

To address these limitations, Hong \etal propose an alternative taxonomy, categorizing domain shifts into covariate, semantic, and contextual shifts.
While this refinement offers greater precision, it does not account for certain semantic shifts, such as those observed across species in benchmarks like \acs{MIDOG} and \acl{PHAKIR} (\cref{sec:datasets}).
Consequently, this work adopts the established taxonomy of \cite{yang2022openood,zhang2024openood} with a key modification: datasets with contextual differences from the \ac{ID} data are classified as \ac{nOOD} if these differences are substantial enough to reasonably anticipate unreliable classification performance.
This approach strikes the balance between the established taxonomy and the need for practical applicability in diverse medical imaging contexts.
\section{OOD: Benchmarks and Metrics}
\label{sec:datasets}
To ensure reliable performance in machine learning models, robust control over their inputs is essential.
\Ac{OOD} detection methods are designed to identify inputs that deviate significantly from a model's training distribution, as such unfamiliar inputs can lead to unreliable predictions.
The primary objective of \ac{OOD} detection is to flag instances that fall outside the data distribution used during training.
However, many existing \ac{OOD} detection methods rely heavily on classification predictions, necessitating a highly effective underlying classification model for each \ac{ID} dataset.
The following subsection introduces the classification tasks and provides an overview of the datasets associated with them, highlighting their relevance to the evaluation of \ac{OOD} detection approaches.

\subsection{Medical Imaging Benchmarks}
\label{ssec:benchmarks}
This work builds upon the foundation of the existing OpenOOD benchmark~\cite{yang2022openood, yang2023full, yang2024generalized, zhang2024openood}, which was originally developed for natural images. 
While following a similar structure and taxonomy, we extend this framework to the medical imaging domain by introducing \ac{OBOOM}.
To achieve this, we selected three datasets from diverse medical imaging fields as \ac{ID} and curated corresponding \ac{csID}, \ac{nOOD}, and \ac{fOOD} datasets (\cref{fig:datasets}). 
To ensure robust evaluation, all \ac{OOD} datasets were further divided into validation and test sets. 
The validation sets are used for hyperparameter tuning, while the test sets were exclusively reserved for assessing \ac{OOD} detection performance.
Additional information on all datasets and their respective splits can be found in \cref{sec:supplementary:datasets}.

\paragraph{MIDOG}
\label{sec:datasets:midog}
The \acl{MIDOG} dataset~\cite{aubreville2023comprehensive} by Aubreville \etal comprises 503 images of \acl{HE} stained histological whole slides.
These images are organized into ten distinct domains, labeled as $\text{1}_{\text{a--c}}$, 2--5, $\text{6}_{\text{a,b}}$, and 7 (\cref{sec:supplementary:datasets:midog}), each exhibiting covariate shifts, semantic shifts, or both. 
These shifts arise from variations introduced by differing imaging hardware and staining protocols, as well as seven different cancer types from both human and canine species.

The dataset contains annotations for mitotic cells and \acp{imposter} in the form of \num{50}\,\(\times\)\,\SI{50}{\px} regions.
To adapt the original object detection challenge into a three-class classification task, we utilized the provided annotations to create separate crops for mitotic cells and \ac{imposter} cells.
Moreover, we created additional crops by shifting each annotation window \SI{100}{\px} to the right, such that these crops do not belong to either of the original categories.
These additional crops were subsequently used as training data for the third class.
To ensure the quality and consistency of the dataset, crops smaller than \num{50}\,\(\times\)\,\SI{50}{\px}, typically occurring at the image borders, were excluded.
Further, all additional crops overlapping with any annotation were also removed. 
The same procedure was applied to the remaining \ac{MIDOG} dataset domains as well.

Since the primary distinction between the \ac{ID} data and the data from domains $\text{1}_\text{b}$ and $\text{1}_\text{c}$ lies in the imaging devices used, we designated them as \ac{csID}.

The remaining domains of the \ac{MIDOG} dataset were classified as \ac{nOOD} because they exhibit semantic shifts and, in most cases, additional covariate shifts.

For the \ac{fOOD} category, we incorporated two datasets associated with distinct medical applications.
(1) \textbf{CCAgT dataset}~\cite{amorim2020novel, atkinson_amorim_ccagt_2022}: This dataset contains images of cervical cancer cells stained using the AgNOR technique, which specifically highlights regions within the cell nuclei.
We utilized the available annotations of cell nuclei and extracted \num{50}\,\(\times\)\,\SI{50}{\px} crops centered on each nucleus.
(2) \textbf{\Acs{FNAC} 2019 dataset}~\cite{saikia2019comparative}: This dataset features Pap-stained images of breast \acs{FNAC} samples. 
The crop extraction involved a binary threshold segmentation step, followed by a series of morphological operations to isolate cell clusters.
From the processed images, the ten largest clusters were identified, and \num{50}\,\(\times\)\,\SI{50}{\px} crops were extracted at the centroids of these clusters.

\paragraph{\acl{PHAKIR}}
\label{sec:datasets:phakir}
The \ac{PHAKIR}-Challenge dataset, introduced by Rueckert \etal~\cite{rueckert2024miccai}, consists of eight publicly available real-world human cholecystectomy videos collected from three hospitals. 
It includes annotations for instance segmentation and keypoint detection of 19 distinct surgical instruments, as well as for intervention phases.
To maintain consistency and to avoid introducing unintended covariate shifts, only the first six videos, containing a subset of ten surgical instruments and originating from a single hospital, are included in this study, while two videos from other hospitals are excluded. 
For this benchmark, we train a classification model to identify surgical instruments. 
Using the provided annotations, we selected 2769 frames containing only a single surgical instrument to simplify the classification task. 
Further, 1891 images that do not contain any instruments are also selected and classified under the category "No-Instrument".

To create separate \ac{ID} and \ac{csID} datasets from the \ac{PHAKIR}-Challenge data, extracted frames were categorized into three smoke levels -- none, medium, and heavy -- based on the annotations from~\cite{rueckert2023smoke}, reflecting the extent of smoke produced by tissue coagulation.
Frames with no visible smoke were designated as the \ac{ID} dataset, while frames with medium and heavy smoke were assigned to the \ac{csID} datasets.
This separation ensures a clear distinction between the \ac{ID} and the \ac{csID} datasets, enabling a robust evaluation of \ac{OOD} detection performance with respect to covariate shifts caused by varying levels of smoke.

The following datasets were selected as \ac{nOOD} datasets, each introducing progressively greater semantic and covariate shifts.
(1) \textbf{\acl{cholec} dataset}~\cite{twinanda2016endonet}: Although also featuring cholecystectomies, the instruments used differ from those in the \ac{ID} dataset. 
Only those frames containing a single instrument were selected.
Additionally, strong black vignettes present in almost all videos were removed by cropping a rectangular region within the vignette, preserving the original aspect ratio of the \ac{ID} images.
(2) \textbf{\Ac{EndoSeg15}}~\cite{bodenstedt2018comparative}: From the \ac{EndoVis} 2015, this dataset includes frames from colorectal surgeries.
We utilized images depicting rigid instruments that were distinct from those included in the \ac{ID} dataset.
(3)~\textbf{\Ac{EndoSeg18}}~\cite{allan20202018}: As part of the \ac{EndoVis} Challenge 2018, this dataset features robotic instruments used in porcine nephrectomy procedures.
To reduce the computational load, we utilized only the official test split from this dataset.
Further, only the left frames from the stereo camera setup were used to maintain consistency with the other datasets.
In contrast to the \ac{ID} and \ac{cholec} datasets, we did not discard frames containing multiple instruments in \ac{EndoSeg15} and \ac{EndoSeg18}.

For \ac{fOOD} evaluation, two datasets from entirely different medical fields were utilized:
(1) \textbf{\acl{KVASIR} dataset}~\cite{jha2020kvasir}: This dataset consists of \num{1000} endoscopic images of polyps inside the bowel. 
(2) \textbf{CATARACTS dataset}~\cite{al2019cataracts}: Containing microscope videos of cataract surgeries, this dataset features a variety of ophthalmological instruments entirely different from those in the \ac{ID} dataset.

\paragraph{\ac{OASIS3}}
\label{sec:datasets:oasis}
The \ac{OASIS3}~\cite{lamontagne2019oasis} dataset comprises \num{2842} primarily high-resolution longitudinal \ac{MRI} and low-dose \ac{CT} scans from \num{1378} subjects, covering modalities such as \ac{T1w}, \ac{T2w}, and FLAIR.
The dataset includes individuals ranging from \ac{CN} to various stages of cognitive decline, up to \acf{AD}.
For this work, we specifically selected \ac{CN} and \ac{AD} subjects to facilitate a clearer distinction in brain tissue morphology.
To ensure high-quality labels, we included only those \ac{MRI} scans for which a clinical diagnosis is available within a 365-day window prior to the \ac{MRI} acquisition date or a 182-day window after.
If no diagnosis was available within this constraints, the last one to three prior diagnoses were reviewed and accepted, provided they consistently indicated \ac{AD}.
All selected scans were preprocessed by resampling to an isotropic voxel spacing of \SI{1}{\cubic\milli\meter} and applying skull-stripping using HD-BET~\cite{isensee2019automated}.
For the \ac{ID} dataset, only \ac{T1w} \ac{MRI} scans were used to train a classification model to distinguish between \ac{CN} and \ac{AD} subjects. 
To evaluate the effects of covariate shifts, we created two \ac{csID} datasets:
(1) \textbf{Modality-based covariate shift}: The first set comprises \ac{T2w} \ac{MRI} scans, introducing a covariate shift due to a change in imaging modality.
(2): \textbf{Scanner-based covariate shift}: All \ac{T1w} scans acquired using a Siemens Vision device were withheld to form the second \ac{csID} dataset, representing a shift in the \ac{MRI} scanner.

The \ac{nOOD} category includes datasets that introduce varying degrees of domain shifts, ranging from subtle to more pronounced differences:
(1) \textbf{\acl{ATLAS} dataset}~\cite{liew2022large}: This dataset introduces a semantic shift as it contains images with stroke lesions, which differ from the \ac{CN} and \ac{AD} patterns in the \ac{ID} dataset. 
Only \ac{T1w} scans were selected, resampled, and skull-stripped to ensure consistency with the \ac{ID} image properties. 
We excluded lower-quality scans to avoid introducing unintended covariate shifts (\cref{sec:supplementary:datasets:oasis}).
(2) \textbf{\acl{BRATS} dataset} \cite{baid2021rsna, menze2014multimodal, bakas2017advancing}: Featuring large gliomas that affect substantial regions of the brain, this dataset presents a more significant semantic shift compared to \ac{ATLAS}. 
As with \ac{ATLAS}, only \ac{T1w} images were included, and no preprocessing was required.
(3) \textbf{\ac{CT} scans from \ac{OASIS3}}: These scans represent a unique \ac{nOOD} category. 
Although they depict the same subjects as the \ac{MRI} scans, technically retaining the \ac{CN} and \ac{AD} labels, the acquisition method differs fundamentally.
While \ac{CT} scans are more sensitive to calcifications~\cite{zhu2008magnetic}, atrophy is less visible~\cite{levy1990efficacy}, with both being relevant for the classification task. 
These differences can result in extracted features that may not align with those from \ac{MRI} scans.
To ensure only regions containing semantic shifts are evaluated, we discarded scans for both the \acl{ATLAS} and the \acl{BRATS} datasets where the pathological area lies outside the region of interest -- a \SI{128}{\cubic\mm} volume centered in the brain.

For the \ac{fOOD} setting, two datasets from entirely different anatomical regions were selected to introduce significant semantic and structural differences:
(1) \textbf{\ac{MSD-H} dataset} \cite{tobon2015benchmark, simpson2019large, antonelli2022medical}: The \ac{MSD-H} dataset includes high-resolution contrast-enhanced \ac{MRI} scans of the heart captured during a single cardiac phase. 
(2) \textbf{\acl{CHAOS} dataset} \cite{CHAOSdata2019, CHAOS2021}: We utilize the in-phase sequences from this dataset, which contains dual-echo \ac{MRI} scans of the abdomen.

\subsection{Metrics}
\label{ssec:metrics}
We adopt the metrics established in the original OpenOOD framework \cite{yang2022openood,zhang2024openood}, treating \ac{OOD} samples as the positive class and use the \ac{AUROC} as the primary metric to evaluate performance across all thresholds.
It assesses the trade-off between the \ac{TPR} and \ac{FPR}, measuring the model's ability to separate \ac{ID} from \ac{OOD}.
In addition, we employ the \ac{FPR}@95 metric, measuring the \ac{FPR} when 95\,\% of the \ac{ID} data is correctly classified.

We also report the \ac{AUPR} two variants: (1) with \ac{ID} samples as the positive class (\acl{AUPRIN}), to measure the model's capacity to identify \ac{ID} data; (2) with \ac{OOD} samples as the positive class (\acl{AUPROUT}), to assess the model’s ability to detect unexpected inputs.
Both \ac{AUPR} metrics compute precision and recall across all classification thresholds, similar to \ac{AUROC}.
While we present all metric results in the supplementary (\cref{sec:supplementary:experiments:results}), we report the harmonic mean of \acl{AUPRIN} and \acl{AUPROUT} as a single, balanced metric for our main results. 
To better assess model performance in imbalanced settings, we replace OpenOOD's~\cite{yang2022openood, zhang2024openood} accuracy metric with the F1-Score. 
Further details on the metric selection can be found in \cref{sec:supplementary:datasets:metrics}.
\begin{table*}[htbp]
    \caption{
    Results of all evaluated \acs{OOD} detection methods, ranked by their average performance (\acs{AUROC}) across all \acs{MIB} groups for the \textbf{\acs{nOOD}} category.
    Additionally, the table includes the averaged \acs{nOOD} performance based on the harmonic mean of \acl{AUPRIN} and \acl{AUPROUT}, as well as the averaged \acs{FPR}@95 metrics. The F1-Score indicates the performance on the \acs{ID} test set.
    Rows are color-coded to indicate the source of information used for \acs{OOD} detection: blue for probabilities or logits, orange for feature space, and green for a combination of both.
    *: The results of the MDSEns method are potentially misleading, as discussed in the main text (\cref{par:top-performing}).}
    \label{tab:main_results}
\centering
{\renewcommand\baselinestretch{1.0}\selectfont\resizebox{\textwidth}{!}{
 \begin{tabular}{l@{\hspace{13pt}}
    >{\centering\arraybackslash\color{dark_gr}}m{1.48cm}>{\centering\arraybackslash}m{1.48cm}>{\centering\arraybackslash\color{dark_gr}}m{1.48cm}@{\hspace{13pt}}
     >{\centering\arraybackslash\color{dark_gr}}m{1.48cm}>{\centering\arraybackslash}m{1.48cm}>{\centering\arraybackslash\color{dark_gr}}m{1.48cm}@{\hspace{13pt}}
     >{\centering\arraybackslash\color{dark_gr}}m{1.48cm}>{\centering\arraybackslash}m{1.48cm}>{\centering\arraybackslash\color{dark_gr}}m{1.48cm}@{\hspace{13pt}}
    c>{\color{dark_gr}}c>{\color{dark_gr}}c} 
        \toprule
        \multirow{2}{*}{} 
         & \multicolumn{3}{c}{\textbf{\acs{MIDOG}} @ F1-Score: \num{81.88}} 
         & \multicolumn{3}{c}{\textbf{\acs{PHAKIR}} @ F1-Score: \num{80.08}} 
         & \multicolumn{3}{c}{\textbf{\acs{OASIS3}} @ F1-Score: \num{73.65}} 
         & \multicolumn{3}{c}{\textbf{Averages for \acs{nOOD}}} \\
        
        & \acs{csID} & \tikzmark{midog_topLeft}\textcolor{black}{\acs{nOOD}} & \acs{fOOD} & \acs{csID} & \tikzmark{phakir_topLeft}\textcolor{black}{\acs{nOOD}} & \acs{fOOD} & \acs{csID} & \tikzmark{oasis_topLeft}\textcolor{black}{\acs{nOOD}} & \acs{fOOD} & \tikzmark{avg_topLeft}$\overline{\text{\acs{AUROC}}}$\,\raisebox{0.3ex}{$\uparrow$} & 
        $\overline{\text{\acs{AUPR}}}$\,\raisebox{0.3ex}{$\uparrow$} & $\overline{\text{\acs{FPR}@95}}$\,\raisebox{0.3ex}{$\downarrow$}\\
    \midrule
\rowcolor{Feature_light} MDSEns*~\cite{lee2018simple_MDS}	& 99.19 & 91.84 & 100.00	& 65.05 & 97.11 & 98.50	& 100.00 & 99.46 & 100.00 & 96.14 & 91.86 & 11.97 \\
\rowcolor{Combined_light} ViM~\cite{wang2022vim_ViM_Residual}	& 59.73 & 62.67 & 84.78	& 72.39 & 81.14 & 55.34	& 98.82 & 98.40 & 100.00 & 80.74 & 70.21 & 48.79 \\
\rowcolor{Feature_light} Residual~\cite{wang2022vim_ViM_Residual}	& 60.26 & 65.78 & 92.35	& 57.12 & 76.99 & 57.31	& 96.97 & 96.70 & 100.00 & 79.82 & 70.18 & 49.94 \\
\rowcolor{Feature_light} MDS~\cite{lee2018simple_MDS}	& 58.60 & 63.21 & 90.91	& 56.04 & 76.48 & 51.47	& 96.31 & 96.15 & 100.00 & 78.61 & 69.72 & 51.29 \\
\rowcolor{Feature_light} KNN~\cite{sun2022out_KNN}	& 57.97 & 61.63 & 90.18	& 34.05 & 55.44 & 37.75	& 98.78 & 97.66 & 100.00 & 71.58 & 61.94 & 62.62 \\
\rowcolor{Feature_light} SHE~\cite{zhang2022out_SHE}	& 57.89 & 61.80 & 91.09	& 36.18 & 50.34 & 47.40	& 91.85 & 90.80 & 99.64 & 67.65 & 58.40 & 67.29 \\
\rowcolor{Feature_light} RMDS~\cite{ren2021simple_RMDS}	& 49.46 & 52.23 & 60.68	& 38.22 & 67.73 & 35.49	& 58.00 & 71.69 & 99.89 & 63.88 & 48.58 & 79.88 \\
\rowcolor{Combined_light} Relation~\cite{kim2024neural_Relation}	& 55.68 & 58.71 & 86.17	& 31.06 & 61.02 & 30.72	& 45.48 & 66.20 & 92.91 & 61.98 & 47.95 & 78.62 \\
\rowcolor{Combined_light} fDBD~\cite{liu2024_fdbd}	& 52.54 & 58.33 & 83.03	& 30.08 & 50.13 & 27.54	& 58.62 & 75.31 & 92.93 & 61.26 & 50.33 & 76.55 \\
\rowcolor{Combined_light} SCALE~\cite{xu2023scaling_SCALE}	& 53.44 & 55.71 & 82.03	& 35.91 & 38.97 & 46.83	& 84.53 & 84.32 & 98.23 & 59.67 & 48.91 & 78.67 \\
\rowcolor{Combined_light} ReAct~\cite{sun2021react_ReAct}	& 53.79 & 57.40 & 84.86	& 30.34 & 48.38 & 25.89	& 44.74 & 70.23 & 77.53 & 58.67 & 46.49 & 81.79 \\
\rowcolor{Combined_light} ASH~\cite{djurisic2022extremely_ASH}	& 53.93 & 56.02 & 82.70	& 45.99 & 40.17 & 65.08	& 70.71 & 76.52 & 91.48 & 57.57 & 46.13 & 80.71 \\
\rowcolor{Combined_light} RankFeat~\cite{song2022rankfeat_RankFeat}	& 48.17 & 51.83 & 56.44	& 52.01 & 45.26 & 27.35	& 71.29 & 74.65 & 96.99 & 57.25 & 48.14 & 80.36 \\
\rowcolor{Classification_light} OpenMax~\cite{bendale2016towards_OpenMax}	& 49.88 & 52.75 & 65.87	& 31.90 & 66.03 & 33.56	& 43.69 & 52.60 & 70.47 & 57.13 & 40.39 & 90.03 \\
\rowcolor{Classification_light} ODIN~\cite{liang2017enhancing_ODIN}	& 56.38 & 61.37 & 85.08	& 34.28 & 41.78 & 71.82	& 53.51 & 59.53 & 58.63 & 54.23 & 44.15 & 86.91 \\
\rowcolor{Classification_light} GEN~\cite{liu2023gen_GEN}	& 52.54 & 55.46 & 79.78	& 32.65 & 51.55 & 32.68	& 36.95 & 53.50 & 78.45 & 53.50 & 37.50 & 91.64 \\
\rowcolor{Classification_light} MSP~\cite{hendrycks2016baseline_MSP}	& 53.37 & 55.90 & 80.91	& 32.04 & 50.16 & 32.51	& 36.95 & 53.50 & 78.45 & 53.19 & 37.29 & 91.71 \\
\rowcolor{Classification_light} Dropout~\cite{gal2016uncertainty_Dropout}	& 53.25 & 55.76 & 80.87	& 31.96 & 50.10 & 32.56	& 37.10 & 53.22 & 77.13 & 53.03 & 37.17 & 91.88 \\
\rowcolor{Classification_light} TempScale~\cite{guo2017calibration_TempScale}	& 53.57 & 56.05 & 81.38	& 31.73 & 48.73 & 32.50	& 36.95 & 53.50 & 78.45 & 52.76 & 37.05 & 91.65 \\
\rowcolor{Combined_light} NNGuide~\cite{park2023nearest_NNGuide}	& 58.08 & 59.84 & 89.58	& 29.15 & 37.98 & 40.94	& 34.76 & 56.95 & 76.37 & 51.59 & 37.08 & 90.55 \\
\rowcolor{Classification_light} KLM~\cite{hendrycks2019benchmark_KLM}	& 48.34 & 51.54 & 73.53	& 55.15 & 56.34 & 35.87	& 54.51 & 44.49 & 63.98 & 50.79 & 36.57 & 89.40 \\
\rowcolor{Classification_light} EBO~\cite{liu2020energy_EBO}	& 54.46 & 56.85 & 84.45	& 31.99 & 40.18 & 34.34	& 33.86 & 49.39 & 69.65 & 48.81 & 35.23 & 92.33 \\
\rowcolor{Classification_light} MLS~\cite{hendrycks2019scaling_MLS}	& 54.22 & 56.58 & 82.98	& 32.00 & 40.17 & 34.33	& 33.94 & 49.47 & 69.73 & 48.74 & 35.22 & 92.37 \\
\rowcolor{Combined_light} DICE~\cite{sun2022dice_DICE}	& 51.56 & 54.15 & 79.08	& 35.96 & 53.33 & 22.65	& 34.66 & 26.14 & 23.70 & 44.54 & 34.22 & 92.83 \\
\bottomrule
    \end{tabular}}\par}
\end{table*}
\section{Experiments and Results}
\label{sec:experiments}
This section begins by outlining the training pipeline for the classification models used with the three \acp{MIB}: \ac{MIDOG}, \ac{PHAKIR}, and \ac{OASIS3}.
Subsequently, the conducted experiments are described and discussed in detail.

\subsection{Classifier Training}
\label{sec:experiments:classifier_training}
All \ac{ID} datasets are split into training, validation, and test subsets.
To optimize the three \ac{MIB}-specific classification models, we use the OneCycle learning rate scheduler~\cite{smith2019super} with either the \acs{SGD} optimizer~\cite{sutskever2013importance} for \ac{MIDOG} and the Adam optimizer~\cite{kingma2014adam} for \ac{PHAKIR} and \ac{OASIS3}. 
Final models, trained using \acl{CE} loss, are selected based on the highest macro-averaged F1-score on the respective validation sets.
To enhance performance and reduce training time, we initialize models with pre-trained weights without freezing any model parameters: \acs{IN1k}~\cite{deng2009imagenet} for \ac{MIDOG} and \ac{PHAKIR}, and Kinetics400~\cite{kay2017kinetics} for \ac{OASIS3}.
\cref{sec:supplementary:experiments:classifier} provides further details on each training setup.

\subsection{Experimental Setup}
\label{sec:experiments:setup}
We evaluate \num{24} post-hoc methods, initially evaluated on natural images, in the context of our \acp{MIB}, with results summarized in \cref{tab:main_results}. 
The methods are ranked based on their average \ac{AUROC} performance across all \acp{MIB} in the \ac{nOOD} setting, as this represents critical, high-risk scenarios where failure in \ac{OOD} detection could lead to potential misdiagnoses or compromised patient safety.

We categorize the evaluated methods into three groups:
(1) \textbf{Classification-Based Methods (blue)}: These methods primarily use information derived from the model output probabilities or logits.
(2) \textbf{Feature-Based Methods (orange)}: These methods rely exclusively on information from the feature space.
(3) \textbf{Hybrid Methods (green)}: These methods combine both sources of information.

For this evaluation, we extend the codebase provided by the OpenOOD benchmark~\cite{zhang2024openood}.
To enable hyperparameter tuning, we aggregate validation splits from all \ac{nOOD} datasets within each benchmark. 
Fine-tuning is conducted on this combined validation set, focusing on optimizing performance in the challenging \ac{nOOD} context.

\input{figures/top4_confidence_scores}
\subsection{Results on all \aclp{MIB}}
\label{sec:experiments:results}
In this subsection we start with a description of the four top-performing methods, followed by a detailed analysis of the challenges posed by specific datasets to those methods.
It also addresses the suboptimal performance of classification-based methods and compares the results with those obtained on the \ac{IN1k} benchmark from OpenOOD~\cite{yang2022openood}.
Additionally, \Cref{tab:method_descriptions_classification} -- \cref{tab:method_descriptions_hybrid} in the appendix provide concise summaries of all the approaches.
\paragraph{MDSEns and MDS}
The MDSEns method~\cite{lee2018simple_MDS} from Lee \etal achieved the best overall performance across all \acp{MIB}.
This method calculates the Mahalanobis distance between feature embeddings of test samples and precomputed class-conditional Gaussian distributions derived from \ac{ID} training data across multiple model layers.
MDSEns aggregates these distance calculations from all intermediate layers through weighted averaging, with weights optimized during an initial setup phase using logistic regression on validation data containing both \ac{ID} and \ac{OOD} examples.

In contrast to MDSEns, the fourth-ranked method MDS~\cite{lee2018simple_MDS} computes the Mahalanobis distance using feature embeddings derived solely from the penultimate layer.

\paragraph{Residual and ViM}
Similar to MDS, the third-ranked method Residual from Wang \etal~\cite{wang2022vim_ViM_Residual} relies entirely on features extracted from the penultimate layer.
The method projects features onto a low-variance subspace defined by the \emph{N} smallest eigenvalues of the empirical covariance matrix calculated from all \ac{ID} training data, and uses the norm of those features as the \ac{OOD} score.

ViM, an extension of the Residual method, integrates class-specific logit information to enhance \ac{OOD} detection~\cite{wang2022vim_ViM_Residual}.
It achieves this by transforming the Residual output into a virtual logit, which is then combined with Energy-Based \ac{OOD} Detection (EBO)~\cite{liu2020energy_EBO}. 
This approach recovers information lost in the feature-to-logit mapping, consistently improving detection performance across diverse \ac{OOD} scenarios with natural images.

\paragraph{Results on challenging datasets}
\label{par:top-performing}
\Cref{fig:top4_confidence_scores} illustrates the distribution of \ac{OOD} scores and corresponding \ac{AUROC} values for the four top-performing methods on particularly challenging datasets from \ac{MIDOG} and \ac{PHAKIR}.
A corresponding visualization for \ac{OASIS3} is presented in \cref{fig:top4_confidence_scores_oasis} in the appendix.
From \cref{tab:main_results} it is evident that MDSEns delivers strong overall performance across the evaluated \acp{MIB}.
However, even this top-performing method faces difficulties in reliably distinguishing between \ac{ID} and \ac{OOD} data under certain challenging conditions, such as given by \ac{MIDOG}'s domain 5 dataset and \ac{PHAKIR}'s Medium Smoke dataset.

This outcome aligns with the characteristics of domain 5 from \ac{MIDOG}, which originates from the same scanner and institution as the \ac{ID} data.
Here, the primary shift is semantic, involving a different cell type from the same species.
In contrast, domain~$\text{6}_\text{a}$ exhibits both a more pronounced semantic shift -- a different cell type from another species -- and additional covariate shifts introduced by variations in scanner type and institution. 
While MDSEns achieves near-perfect separation in domain~$\text{6}_\text{a}$, where both semantic and covariate shifts are present, it degrades significantly in domain 5, where only a semantic shift is involved. 

This finding indicates that MDSEns' strong performance is heavily influenced by its ability to detect covariate shifts, which are often more prominent in the datasets used.
MDSEns uniquely uses information from early intermediate layers of the network, where covariate shifts are more likely to manifest as changes in basic features like edges, shapes, and colors~\cite{krizhevsky2012imagenet, zeiler_visualizing}.
This observation aligns with prior findings~\cite{vasiliuk2023limitations}, demonstrating that simple approaches based on input intensity distributions can yield results comparable to methods that rely on deeper network layers or final outputs.

For the \ac{PHAKIR} Medium Smoke dataset, the overlap between \ac{OOD} and \ac{ID} scores can be attributed to the subtle appearance of smoke in the images. 
Often, the smoke is limited to regions that do not contain instruments.
Given that the model's primary task is instrument classification, it likely prioritizes regions containing instruments over background areas.
As a result, features associated with smoke, particularly in non-instrument regions, are minimally encoded.
A representative example of this behavior is shown in \cref{fig:attribution} in the appendix.
If the downstream task were instead organ classification, it is likely that smoke would have a stronger impact on the encoded features, as it could obscure or alter key characteristics of the organs themselves. 

Comparing the performance of Residual and ViM~\cite{wang2022vim_ViM_Residual} on \ac{PHAKIR}'s \ac{csID} Medium Smoke dataset reveals significant performance degradation (\cref{fig:top4_confidence_scores}) for ViM.
While Residual performs robustly in this scenario, ViM, which combines Residual with the EBO  method~\cite{liu2020energy_EBO}, exhibits a noticeable drop in performance. 
This degradation is explained by EBO's poor performance on this dataset, with an \ac{AUROC} of \num{23.74}. 

\paragraph{Poor performance of classification-based methods}
An evaluation of the average \ac{AUROC} for classification-based approaches across the three \acp{MIB} reveals a significant lack of discriminative power compared to feature-based methods (\cref{tab:average_results} in the appendix).
These approaches inherently rely on the assumption that \ac{ID} data should yield higher confidence scores than \ac{OOD} data. 
However, this assumption is often violated in practice, as demonstrated by our analysis.

For example, in the \ac{PHAKIR} benchmark, the mean softmax probability for \ac{ID} predictions of the "Grasper" class is \num{79.07}\,\%, whereas for \ac{OOD} predictions from the \ac{EndoSeg18} dataset, the mean probability on all Grasper predictions increases to \num{94.04}\,\%.
This paradoxical result indicates a critical flaw: the classifier overconfidently assigns the Grasper label to \ac{OOD} data. 
Analyzing the feature space reveals that \ac{OOD} data from \ac{EndoSeg18} clusters predominantly near the Grasper class (\cref{fig:phakir_tsne} in the appendix), suggesting a bias in the classifier toward assigning this label to instruments that do not clearly belong to any other class. 
Together, these factors explain the low average \ac{AUROC} of \num{33.71} for classification-based methods on the \ac{EndoSeg18} dataset.

Similar patterns of poor discrimination are observed across other datasets.
In the \ac{KVASIR} dataset, for example, the classifier assigned higher scores to predictions for the "No-Instrument" class than it did for \ac{ID} data.
Similarly, in the \ac{OASIS3} benchmark, outputs for the \ac{CN} class are nearly identical between the \ac{ID} and the \ac{ATLAS} dataset, compromising the ability of classification-based methods to distinguish between \ac{ID} and \ac{OOD} samples in each case.

\paragraph{Comparison with OpenOOD's \ac{IN1k} benchmark}
\input{figures/mib_vs_in}
\Cref{fig:benchmarks_vs_imagenet} compares the average \ac{AUROC} across the  \acp{MIB} for all evaluated methods with their performance on the \acl{IN1k} benchmark from OpenOOD~\cite{yang2022openood}. 
The results highlight that no single method fully addresses the challenge of \ac{OOD} detection across both natural and medical imaging scenarios.
The closest contender of a comprehensive approach is ViM, which combines information from both feature space and the classification layers.
ViM achieves an average \ac{AUROC} of 80.74 on the \acp{MIB} and 72.08 on the \ac{IN1k} benchmark, demonstrating its adaptability to different domains.
However, the performance gap between these benchmarks underscores the complexity and specificity of \ac{OOD} detection in medical imaging, where domain characteristics significantly differ from those of natural images.
Notably, methods that rely primarily on feature space information tend to perform better on medical images, while methods leveraging probabilities or logits show superior results on natural images. 
This discrepancy may stem from the lower variance typically observed in medical images compared to natural images.
For instance, the average standard deviation in pixel intensity is \num{0.148}/\num{0.149} for the \ac{MIDOG}/\ac{PHAKIR} \ac{ID} datasets, respectively, compared to \num{0.226} for the \ac{IN1k} dataset.
The lower variance in medical images, along with the typically small number of target classes in medical datasets, may result in more cohesive and less fragmented feature spaces, thereby enhancing the effectiveness of feature-based methods for detecting \ac{OOD} samples, presenting a starting point for future research.
On the other hand, natural images, with their higher variability and richer textures, likely benefit from methods that exploit class probabilities and logits.
Detailed results on \acl{IN1k} can be found in \cref{sec:supplementary:experiments:results}.%
\section{Conclusion}
\label{sec:conclusion}
In this work, we introduced the \acf{OBOOM}, representing a significant step forward in the evaluation and development of \ac{OOD} detection methods for medical imaging. 
\Ac{OBOOM} comprises a comprehensive set of three benchmarks consisting of 14 datasets, categorized into \acl{ID}, \acl{csID}, \acl{nOOD}, and \ac{fOOD}.
Across these benchmarks, we evaluated 24 post-hoc \ac{OOD} detection methods and analyzed their performance and limitations in the medical field.

Our results reveal that state-of-the-art \ac{OOD} detection methods that rely on feature space information consistently outperform methods that depend solely on probabilities and logits when applied to medical datasets.
However, our findings also highlight a critical challenge: methods originally designed and optimized for natural images often fail to generalize to medical data.
This underscores the need for tailored solutions that address the unique characteristics of medical imaging, such as lower variance, domain-specific semantic shifts, and imbalanced datasets. 
The precise reasons behind these discrepancies remain an area for further research.
One notable limitation of our benchmark is its focus on classification tasks, which limits the applicability of our findings to such problems.
Future work could extend the evaluation framework to seamlessly include segmentation tasks, a critical component of medical imaging, by adapting the included \ac{OOD} detection methods to operate in segmentation settings.
While we focus on evaluating \ac{OOD} detection methods using deep \aclp{NN}, we recognize the potential of other approaches.
For example, analyzing and evaluating input data distributions and directly comparing them to training set distributions could complement \acl{NN}-based \ac{OOD} detection.

As \acl{AI} continues to advance in the medical field, the insights and benchmarks provided by \ac{OBOOM} will play a vital role in driving innovation.
By exposing the strength and weaknesses of existing methods and setting a high standard for future development, \ac{OBOOM} contributes to the creation of trustworthy \ac{AI} systems.
These systems will be essential for safely and effectively addressing the challenges of clinical application, ensuring that \ac{AI} technologies enhance patient care while maintaining high standards of reliability and safety.

\pagebreak

{
    \small
    \bibliographystyle{ieeenat_fullname}
    \bibliography{main}
}

\clearpage
\setcounter{page}{1}
\maketitlesupplementary
\appendix

\section{Datasets}
\label{sec:supplementary:datasets}
This section provides a detailed overview of all datasets used in this work.  
To facilitate reproducibility, we include preprocessing scripts for each dataset in our public GitHub repository, enabling the transformation of the downloaded datasets into the utilized \ac{ID} and \ac{OOD} datasets.  
For clarity, the fundamental steps executed by these scripts are outlined below.

\subsection{\ac{MIDOG} benchmark}
\label{sec:supplementary:datasets:midog}
\paragraph{\ac{MIDOG}~\cite{aubreville2023comprehensive}} 
The \ac{MIDOG} dataset consists of 503 whole slide images stained with \acl{HE}, a widely used stain for differentiating tissue components and evaluating tissue morphology. 
The dataset is divided into ten distinct domains, labeled as 1$_\text{a}$, 1$_\text{b}$, 1$_\text{c}$, 2, 3, 4, 5, 6$_\text{a}$, 6$_\text{b}$, and 7. 
Annotations are provided for mitotic cells and \ac{imposter} cells.  
The preprocessing steps outlined in \cref{sec:datasets:midog} yielded \num{451} mitotic cell crops, \num{724} \ac{imposter} cell crops, and \num{1153} additional crops extracted from \num{50} whole slide images within the \ac{ID} domain $\text{1}_{\text{a}}$.
Each domain's number corresponds to a semantic cell type shift, stemming from seven different cancer types and two species: human and canine. 
The cancer types include breast carcinoma, lung carcinoma, lymphosarcoma, cutaneous mast cell tumor, neuroendocrine tumor, soft tissue sarcoma, and melanoma.
 Furthermore, the domains display differing levels of covariate shift caused by variations in imaging hardware and staining protocols. 
 Subscripts are used to indicate domains with multiple sources of covariate shift. 
 Domains 2, 3, 4, $\text{6}_{\text{a}}$, and $\text{6}_{\text{b}}$ exhibit covariate shifts, whereas domains 5 and 7 do not show apparent covariate shifts, as their images were generated using the same imaging hardware as the \ac{ID} dataset and originate from the same institute, employing the same staining protocol as the \ac{ID} set.
The whole \ac{MIDOG} dataset serves as \ac{ID} (1$_\text{a}$), \ac{csID} (1$_\text{b}$, 1$_\text{c}$), and \ac{nOOD} (2 -- 7) datasets.
\paragraph{CCAgT \cite{amorim2020novel, atkinson_amorim_ccagt_2022}} 
    The CCAgT dataset comprises 15 tissue slides stained using the AgNOR technique, labeled alphabetically from \textquoteleft A\textquoteright\ to \textquoteleft O\textquoteright. 
    The AgNOR stain specifically targets regions within the cell nucleus, providing insights into distinct cellular properties. 
    Nuclei annotations are available for these slides and were used to generate the same type of image crops as those from the various domains of the \ac{ID} dataset (\cref{sec:datasets:midog}). 
    This process produced \num{29675} crops which are utilized as the first \ac{fOOD} dataset.
\paragraph{\ac{FNAC} 2019~\cite{saikia2019comparative}}
    The \ac{FNAC} 2019 dataset comprises 212 images of human breast tissue samples obtained via \acl{FNAC}. 
    Of these, 113 images are classified as malignant, while 99 are labeled as benign. Due to the absence of cell-level annotations, we extract cell crops through a multi-step processing pipeline. 
    First, each image is segmented using a binary threshold with a value of 100. 
    Next, morphological opening with a kernel size of 5 and erosion with a kernel size of 3 are applied to isolate cell clusters. 
    From the resulting processed images, the ten largest clusters are identified, and \num{50}\,\(\times\)\,\SI{50}{\px} crops are generated around the centroids of these clusters. 
    The resulting 2088 crops are subsequently utilized as the second \ac{fOOD} dataset.

\subsection{\ac{PHAKIR} benchmark}
\label{sec:supplementary:datasets:phakir}
\paragraph{Acknowledgements}
We thank the creators of the \ac{PHAKIR} dataset for granting permission to use their dataset ahead of the challenge results' publication.
\begin{table*}[htbp]
\caption{Summary of available frames for each instrument class. Video 06 is employed as test data for the official challenge evaluation and therefore not publicly available.}
    \label{tab:phakir:frame_data}
\centering
{
 \begin{tabular}{lccccccc} 
\toprule              
   & Video 01 & Video 02 & Video 03 & Video 04 & Video 05 & Video 07 & Sum \\
\midrule
    Clip-Applicator &  63 & 151 &  53 &  22 &  26 &   0 &  315 \\
    Grasper         &  40 &  13 &   7 &  81 &  52 & 125 &  318 \\
    PE-Forceps      &  68 & 891 &  72 & 109 &  52 &  42 & 1234 \\
    Needle-Probe    &  20 &  27 &   6 &  29 &  12 &  31 &  125 \\
    Palpation-Probe &  18 &  45 &  35 & 187 &  25 & 110 &  420 \\
    Suction-Rod     &  20 &  96 &   7 &  45 &  37 & 152 &  357 \\
    No-Instrument   & 198 & 483 & 323 & 442 & 166 & 279 & 1891 \\
\bottomrule
    \end{tabular}}
\end{table*}
\paragraph{\ac{PHAKIR}~\cite{rueckert2024miccai}} 
The \ac{PHAKIR} dataset consists of eight endoscopic videos of cholecystectomy procedures, with annotations for 19 instrument classes provided as segmentation masks and keypoints for every 25th frame.
For this study, only frames containing a single surgical instrument were selected. 
However, one instrument class, the trocar, is exclusively an access instrument and, therefore, frequently visible alongside other surgical instruments. 
Consequently, frames showing a trocar in conjunction with a single surgical instrument were also included and assigned the label of the accompanying surgical instrument.

To enhance the object-to-background ratio in the selected frames, frames were excluded if the instrument covered less than \num{0.5}\% of the image area or if the distance between the instrument's endpoint and tip was less than \SI{150}{\px}. 
In the \ac{PHAKIR} dataset, the tip refers to the part of the instrument that directly contacts the organ, while the endpoint denotes the location where the instrument appears at the image border. 
Rueckert \etal~\cite{rueckert2023smoke} provided annotations for the first four videos of the \ac{PHAKIR}-Challenge dataset. 
In this work, we extend these annotations to include Video 05 and Video 07.
Following the previously established filtering process, each frame was categorized into three levels of smoke intensity by utilizing the respective annotations -- none, medium, and heavy -- using the corresponding annotations.
The categorization criteria were as follows: None, if no smoke was perceptible; Medium, if smoke was present but the instrument remained clearly distinguishable; and Heavy, if the instrument was no longer clearly distinguishable.
Frames without visible smoke from the first six videos were designated as \ac{ID} data, while frames containing medium or heavy smoke were used as \ac{csID} data.

Within the \ac{ID} dataset, instrument classes with fewer than \num{80} available training images were excluded, resulting in a final dataset of \num{2769} frames across six instrument classes (\cref{tab:phakir:frame_data}). 
To prevent an unintended semantic shift, images from excluded instrument classes were also removed from the \ac{csID} sets.
    
\paragraph{\acs{cholec}~\cite{twinanda2016endonet}} 
The \acs{cholec} dataset comprises 80 endoscopic videos of cholecystectomies, with annotations identifying instrument classes present in every 25th frame. 
Consistent with the methodology used for \ac{PHAKIR}, only frames containing a single surgical instrument were selected for analysis.

This dataset's role within the \ac{MIB} is to evaluate semantic shifts arising from variations in surgical instruments. 
Consequently, instrument classes that overlap with those in \ac{PHAKIR} were excluded, yielding a total of \num{74049} frames.

Most videos, except for videos 40, 60, 65, and 80, exhibit a pronounced black vignette. 
To avoid introducing unintended covariate shifts, a rectangular region within the vignette was cropped while maintaining the original aspect ratio of the \ac{ID} images. 
The dataset is employed as a \ac{nOOD} dataset.
\paragraph{\acs{EndoSeg15}~\cite{bodenstedt2018comparative}} 
The \ac{EndoSeg15} dataset from the \ac{EndoVis} 2015 challenge comprises 160 training images, evenly distributed across four distinct laparoscopic surgeries. 
Similar to the \ac{cholec} dataset, the frames in \ac{EndoSeg15} often contain black vignettes or borders, which we exclude by extracting rectangular crops with the same aspect ratio as the \ac{ID} dataset.
This crop is carefully positioned within the vignette or usable image content, ensuring the exclusion of black borders. 
The resulting dataset is then used as a \ac{nOOD} dataset.
\paragraph{\acs{EndoSeg18}~\cite{allan20202018}} 
The \ac{EndoSeg18} test dataset, part of the \ac{EndoVis} Challenge 2018, consists of 1000 frames captured during four porcine surgical procedures featuring robotic instruments. 
No preprocessing was applied to this dataset and it was utilized for \ac{nOOD} detection evaluation.
\paragraph{\ac{KVASIR}~\cite{jha2020kvasir}} 
The \ac{KVASIR} dataset, an extension of the original Kvasir dataset introduced by Pogorelov \etal~\cite{pogorelov2017kvasir}, comprises 1000 images of colorectal polyps, from which ten contain surgical instruments different from those in the \ac{ID} dataset.
\ac{KVASIR} is employed as a \ac{fOOD} dataset.
    
\paragraph{CATARACTS~\cite{al2019cataracts}} 
The CATARACTS dataset comprises videos of cataract surgeries and includes 21 ophthalmological instruments, which are completely distinct from those in the \ac{ID} dataset. 
Given the large size of this dataset, we limit our analysis to the first five videos from the official test split.
The initial 43, 203, 130, 29, and 159 frames were excluded from these videos, because these frames contain only black content.
Afterwards, the dataset still contains \num{181986} usable frames. 
CATARACTS is utilized as the second \ac{fOOD} dataset.

\subsection{\ac{OASIS3} benchmark}
\label{sec:supplementary:datasets:oasis}
\paragraph{Acknowledgements}
Data were provided in part by OASIS-3: Longitudinal Multimodal Neuroimaging: Principal Investigators: T. Benzinger, D. Marcus, J. Morris; NIH P30 AG066444, P50 AG00561, P30 NS09857781, P01 AG026276, P01 AG003991, R01 AG043434, UL1 TR000448, R01 EB009352. AV-45 doses were provided by Avid Radiopharmaceuticals, a wholly owned subsidiary of Eli Lilly.

\paragraph{\ac{OASIS3}~\cite{lamontagne2019oasis}} 
The longitudinal \ac{OASIS3} dataset comprises 2842 MRI scans from 1378 subjects, covering multiple modalities, including \ac{T1w} and \ac{T2w} scans. 
Each scan is labeled with the number of days since the subject's initial visit.
Additionally, clinical diagnoses, also timestamped by days since the initial visit, are provided. 
However, these timestamps do not precisely align between the clinical diagnoses and \ac{MRI} scans, requiring the matching of clinical diagnoses to imaging visits as described in Section \ref{sec:datasets:oasis}.

We categorize the dataset into subjects with clinical diagnoses of \acf{CN} and \acf{AD}, using the provided diagnostic labels. 
Subject 30753 was excluded due to the absence of a diagnosis, and subjects 30937 and 31357 were discarded as they lacked \ac{MRI} scans. 
Additionally, \ac{MRI} scans for which the skull-stripping process failed were excluded. 
Specifically, this affected \ac{T2w} scans from subjects 30649, 30724, and 30815, as well as a \ac{T1w} scan from subject 30339.

In cases where multiple \ac{MRI} scans were available for a single acquisition timestamp, the scan with the highest index number was selected. 
\ac{MRI} scans containing only the hippocampal region were excluded to avoid introducing an unintended domain shift.

For each \ac{MRI} scan, we reviewed the associated metadata to identify the acquisition device used. 
One out of eight scanners, namely the \emph{Siemens Vision} device, exhibited a unique orientation and axis configuration that differed from other devices. 
To ensure consistency with the default orientation of the broader dataset and prevent an unintended domain shift, these images were realigned to match the orientation used by other devices.

\ac{T1w} \ac{MRI} scans from all devices, except the Siemens Vision, are designated as \ac{ID} data. 
The corresponding \ac{T2w} \ac{MRI} scans from these subjects, if available, are labeled as \ac{csID} data, due to the change in imaging modality. 
The withheld \ac{T1w} \ac{MRI} scans from the Siemens Vision device are labeled as \ac{csID}, as the covariate shift arises from differences in the acquisition device.

Following these preprocessing steps, the final dataset consists of 944 \ac{CN} and 288 \ac{AD} \ac{MRI} scans, forming the \ac{ID} dataset.

\paragraph{\ac{ATLAS}~\cite{liew2022large}} 
The \ac{ATLAS} challenge dataset consists of 33 cohorts, each containing multiple subjects with brain lesions resulting from strokes. 
For this study, we exclusively utilize the \ac{T1w} \ac{MRI} scans from the official training split. 
Cohorts R027, R047, R049, and R050 were excluded from our analysis due to significant quality degradation compared to the remaining cohorts. 
Consequently, a total of 595 \ac{MRI} scans were included in the analysis as the first \ac{nOOD} dataset.

All selected \ac{OASIS3} and \ac{ATLAS} \ac{MRI} scans are preprocessed by resampling to an isotropic voxel spacing of \SI{1}{\cubic\milli\meter} and applying skull-stripping using HD-BET~\cite{isensee2019automated}, version 2.0.1 (official release).

\paragraph{\ac{BRATS} \cite{baid2021rsna, bakas2017advancing, menze2014multimodal}} The \ac{BRATS} 2023 Glioma challenge dataset includes subjects with large gliomas in the brain. 
As the data was preprocessed prior to release, including steps such as resampling and skull-stripping, no further preprocessing was necessary. 
Therefore, the complete official training split, comprising 1251 \ac{T1w} \ac{MRI} scans, was used as the second \ac{nOOD} dataset.

\paragraph{\acs{CT} from \ac{OASIS3}~\cite{lamontagne2019oasis}}
In addition to the \ac{MRI} data, the \ac{OASIS3} dataset includes 1472 low-dose \ac{CT} scans, which were acquired to perform attenuation correction for PET scans~\cite{lamontagne2019oasis}. 
To emphasize brain tissue, we clipped these scans to a range of \numrange{0}{80} Hounsfield units and subsequently normalized them. 
The resulting dataset was used as the third \ac{nOOD} dataset.

\paragraph{\acs{MSD-H} \cite{antonelli2022medical, tobon2015benchmark}} 
The \ac{MSD-H} dataset consists of 30 \ac{MRI} scans of the human heart, all acquired during a single cardiac phase using a 3D balanced steady-state free precession acquisition method. 
These scans were initially employed in a benchmark for left atrium segmentation~\cite{tobon2015benchmark}. 
The dataset encompasses images of varying quality, ranging from high-resolution scans to those with substantial noise.
Utilized as the first \ac{fOOD} dataset.

\paragraph{\acs{CHAOS} \cite{CHAOSdata2019, CHAOS2021}}
The official test split of the \ac{CHAOS} challenge dataset consists of 20 \ac{MRI} scans of the abdomen, originally designed for the task of segmenting abdominal organs. The dataset includes both in-phase and out-of-phase images from dual-echo \ac{MRI} sequences. 
For the \ac{fOOD} evaluation, we use the in-phase scans, as they exhibit stronger visual alignment with the imaging characteristics of the \ac{ID} dataset, while still maintaining significant anatomical differences.
\subsection{Splits}
\label{sec:supplementary:datasets:splits}
\begin{table*}[tbp!]
\caption{Whole slide image identifiers utilized for test and validation.
1--7 denote the different domains of the \ac{MIDOG} dataset.
The last row describes the total number of extracted crops per domain.}
    \label{tab:midog_test_val_split}
\centering
{\renewcommand\baselinestretch{1.0}\selectfont\resizebox{\linewidth}{!}{
 \begin{tabular}{lccccccccc} 
\toprule              
   & $\text{1}_\text{b}$ & $\text{1}_\text{c}$& 2   &  3  & 4    & 5     & $\text{6}_\text{a}$ &  $\text{6}_\text{b}$    &  7 \\
\midrule
    \textbf{Test} & \numrange{51}{95} & \numrange{101}{145}   &  \numrange{200}{240} &  \numrange{245}{294} & \numrange{300}{344} & \numrange{350}{399} & \numrange{405}{481} & \numrange{490}{503} & \numrange{505}{549} \\
    \textbf{Valid.} & \numrange{96}{100} & \numrange{145}{150} & \numrange{241}{244} & 
    \numrange{295}{299} & \numrange{345}{349} & \numrange{400}{404} & \numrange{482}{489} & 504 & \numrange{550}{553} \\
\midrule
    \# Crops & 3258 & 3174 & 3548 & 16059 & 7202 & 4743 & 6260 & 974 & 4084 \\
\bottomrule
    \end{tabular}}\par}
\end{table*}
\paragraph{MIDOG}
The following randomly selected whole-slide image identifiers from \ac{MIDOG}'s \ac{ID} domain $\text{1}_\text{a}$ were utilized for each respective split:
\begin{itemize}
  \item[] \textbf{Train}: 1, 2, 4, 5, 6, 7, 8, 9, 10, 11, 12, 13, 14, 15, 16, 18, 19, 21, 22, 23, 24, 28, 29, 30, 34, 35, 36, 37, 38, 39, 40, 41, 42, 43, 44, 45, 46, 47, 48, 50
  \item[] \textbf{Validation}: 20, 26, 31, 32, 33
  \item[] \textbf{Test}: 3, 17, 25, 27, 49
\end{itemize}
The remaining splits of the \ac{MIDOG} dataset are detailed in \cref{tab:midog_test_val_split}. 
For the CCAgT dataset, the 15 available slides were randomly divided into three validation slides and twelve test slides, corresponding to a \numrange{20}{80} split.
\begin{itemize}
    \item[] \textbf{Validation}: E, L, O
    \item[] \textbf{Test}: A, B, C, D, F, G, H, I, J, K, M, N
\end{itemize}
The FNAC 2019 dataset comprises images categorized as either benign or malignant, and we partitioned each category separately into 10\,\% validation and 90\,\% test subsets, resulting in:
\begin{itemize}
    \item[] \textbf{Validation}: Benign \numrange{90}{99}; Malignant \numrange{103}{113}
    \item[] \textbf{Test}: Benign \numrange{1}{89}; Malignant \numrange{1}{102}
\end{itemize}

\paragraph{\ac{PHAKIR}}
From the six available \ac{ID} videos, one was randomly selected for validation and another for testing.
\begin{itemize}
  \item[] \textbf{Train}: Video 02, 03, 04, 07
  \item[] \textbf{Validation}: Video 05
  \item[] \textbf{Test}: Video 01
\end{itemize}

Similarly to the \ac{ID} split, we use Video 05 as the validation data for the \ac{csID} datasets Medium Smoke and Heavy Smoke, while the remaining five videos serve as the test datasets. 
For the \ac{cholec} dataset, we employ a \numrange{10}{90} split, resulting in:
\begin{itemize}
    \item[] \textbf{Validation}: Videos \numrange{73}{80}
    \item[] \textbf{Test}: Videos \numrange{1}{72}
\end{itemize}
For \ac{EndoSeg15} and \ac{EndoSeg18}, the first three videos of each dataset are designated as test data, while the remaining one is used for validation. 
Similarly, in the CATARACTS dataset, the first four videos are allocated as test data, with the final video serving as validation data. 

For the \ac{KVASIR} dataset, a \numrange{10}{90} validation--test split is applied. 
Given the dataset's size and the use of filenames resembling globally unique identifiers (GUIDs) individual images, it is impractical to list the exact split.
Therefore, detailed information about the data split can be found in the accompanying public GitHub repository.

\paragraph{\ac{OASIS3}}
After preprocessing all \ac{T1w} and \ac{T2w} \ac{MRI} scans as outlined in \cref{sec:supplementary:datasets:oasis}, the remaining \ac{CN} and \ac{AD} data were randomly divided into 70\,\% for training, 15\,\% for validation, and 15\,\% for testing. 
For the \ac{csID} datasets Modality and Scanner, a randomized \numrange{10}{90} split was applied for validation and test sets.

The \ac{ATLAS} dataset was partitioned by assigning 10\,\% of the \ac{MRI} scans from each cohort to the validation set, with the remaining 90\,\% allocated to the test set. 
For the \ac{MSD-H} dataset, instead of performing a random split, the official test split was utilized as validation data, while the official training split was used as test data.

For the other benchmark datasets, specifically \ac{BRATS}, \ac{CT} data from \ac{OASIS3}, and \ac{CHAOS}, the data were randomly divided into 10\,\% for validation and 90\,\% for testing.

Detailed information on the dataset sizes and specific splits can be found in the associated public GitHub repository, which includes the exact subject identifiers and filenames for each partition.

\subsection{Metrics}
\label{sec:supplementary:datasets:metrics}
\begin{table}[tbp!]
\caption{Table showing the number of \ac{ID} test samples relative to the average number of \ac{csID} and \ac{OOD} samples. The red number indicates the factor by which the average number of \ac{csID} and \ac{OOD} samples exceeds that of the \ac{ID} set.}
    \label{tab:average_imbalance}
\centering
{\renewcommand\baselinestretch{1.0}\selectfont\resizebox{\linewidth}{!}{
 \begin{tabular}{l@{\hspace{13pt}}
    >{\centering\arraybackslash}l
    >{\centering\arraybackslash}l
    >{\centering\arraybackslash}l} 
\toprule              
    Data source  & \ac{MIDOG}     & \ac{PHAKIR}   & \ac{OASIS3}  \\
\midrule
\ac{ID} test & 251 & 427 & 181 \\
$\overline{\text{\ac{csID} \& \ac{OOD}}}$ & 6110 \textcolor{red}{$\times$ 24.34} & \num{25350} \textcolor{red}{$\times$ 59.36} & 595 \textcolor{red}{$\times$ 3.29} \\
\bottomrule
    \end{tabular}}\par}
\end{table}
In this work, we employ the \ac{AUPRIN} and \ac{AUPROUT} metrics to assess \ac{OOD} detection performance. 
However, interpreting these metrics can be challenging due to significant imbalances between \ac{ID} and \ac{OOD} data, which are inherent to many \ac{OOD} detection tasks. 
Specifically, the number of \ac{OOD} samples often greatly exceeds the number of \ac{ID} samples, as shown in \cref{tab:average_imbalance}. 
This imbalance directly influences precision, defined as $\frac{TP}{TP+FP}$. 
A higher number of \ac{OOD} samples increases the likelihood of false positives, leading to lower \acl{AUPRIN} values due to reduced precision for \ac{ID} samples. 
Conversely, \acl{AUPROUT} values are generally higher because the abundance of \ac{OOD} samples skews precision favorably when \ac{OOD} is treated as the positive class. 
Recall, defined as $\frac{TP}{TP+FN}$, is similarly affected by these imbalances. 

This effect is particularly evident in the results from the \ac{cholec} and CATARACTS datasets within the \ac{PHAKIR} benchmark, as presented in \cref{tab:phakir_auroc_fpr}. 

\begin{table*}[tbp!]
\caption{Metadata for the training pipeline of each \ac{MIB} classifier. LR stands for learning rate, WD for weight decay, and BS for batch size.}
\label{tab:classifier_training}
\centering
\small
 \begin{tabularx}{\textwidth}{Xlllccccc}
\toprule              
    \ac{MIB}  & Split & Architecture & Optimizer & Seed & Epochs & LR & WD & BS \\
\midrule
    \multirow{2}{*}{\ac{MIDOG}} & 
    \multirow{2}{*}{80-10-10 randomly} & 
    \multirow{2}{*}{ResNet50~\cite{he2016deep}} & 
    \multirow{2}{*}{\shortstack[l]{SGD with \\$\beta$ of \num{0.9}~\cite{sutskever2013importance}}} & 
    \multirow{2}{*}{\num{0}} & 
    \multirow{2}{*}{\num{300}} & 
    \multirow{2}{*}{\num{5e-4}} & 
    \multirow{2}{*}{\num{1e-1}} & 
    \multirow{2}{*}{\num{128}}  
    \\
    \\
    
    \multirow{2}{*}{\ac{PHAKIR}} & 
    \multirow{2}{*}{\shortstack[l]{6 videos split into\\ 4 train, 1 validation, 1 test}} & 
    \multirow{2}{*}{ResNet18~\cite{he2016deep}} & 
    \multirow{2}{*}{Adam~\cite{kingma2014adam}} & 
    \multirow{2}{*}{\num{0}} & 
    \multirow{2}{*}{\num{500}} & 
    \multirow{2}{*}{\num{1e-4}} & 
    \multirow{2}{*}{\num{1e-3}} & 
    \multirow{2}{*}{\num{48}}  
    \\
    \\
    
    \multirow{2}{*}{\ac{OASIS3}} & 
    \multirow{2}{*}{70-15-15 randomly} & 
    \multirow{2}{*}{R(2+1)D~\cite{tran2018closer}} & 
    \multirow{2}{*}{Adam~\cite{kingma2014adam}} & 
    \multirow{2}{*}{\num{0}} & 
    \multirow{2}{*}{\num{300}} & 
    \multirow{2}{*}{\num{5e-5}} & 
    \multirow{2}{*}{\num{1e-6}} & 
    \multirow{2}{*}{\num{15}}  
    \\
    \\
\bottomrule
\end{tabularx}
\end{table*}
\input{figures/top4_confidence_scores_oasis}
In clinical applications, however, machine learning models are more frequently exposed to \ac{ID} data, where detecting rare occurrences of \ac{OOD} inputs becomes crucial. 
Consequently, developers might prioritize highly sensitive \ac{OOD} detection methods (high \ac{AUPROUT} values) to ensure such inputs are reliably flagged. 
At the same time, it is equally important to minimize false positive \ac{OOD} detections (high \ac{AUPRIN} values), as these can compromise the system's usability. 
To address this trade-off, we report the harmonic mean of \acl{AUPRIN} and \acl{AUPROUT} in the main text (\cref{tab:main_results}), calculated as:
\begin{equation*}
    \text{\ac{AUPR}} = \frac{2 \cdot (\text{\ac{AUPRIN}} \cdot \text{\ac{AUPROUT}})}{\text{\ac{AUPRIN}} + \text{\ac{AUPROUT}}}
\end{equation*}
This approach prevents a weak performance in one metric from being overshadowed by strong results in the other, as can occur with the arithmetic mean.

By reporting a comprehensive set of metrics, including \ac{AUROC} (overall \ac{OOD} detection performance), \ac{FPR}@95 (threshold-specific behavior), and \ac{AUPRIN}/\ac{AUPROUT} (detailed insights into \ac{ID} and \ac{OOD} detection performance), we provide a nuanced evaluation of \ac{OOD} detection performance. 

\section{Experiments}
\label{sec:supplementary:experiments}

\subsection{Classifier Training}
\label{sec:supplementary:experiments:classifier}
\Cref{tab:classifier_training} provides additional information regarding the training of each benchmark classifier. To be able to reuse existing model architectures, the final fully connected layer is replaced with a new one, where the output dimension corresponds to the number of classes in each respective classification task.

For training the \ac{MIDOG} and \ac{PHAKIR} classifiers, the \ac{CE} loss function was weighted according to the inverse distribution of class frequencies. 
However, the \ac{PHAKIR} training data was highly imbalanced, particularly with respect to the PE-Forceps class, which was overrepresented due to 891 images from Video 02 (\cref{tab:phakir:frame_data}), as well as the overall dominance of the No-Instrument class. 
To address this imbalance and stabilize the training process, 200 images from the PE-Forceps class in Video 02 and 400 images from all \ac{ID} training videos were randomly sampled in each epoch, with the remaining images withheld for that epoch.

A similar imbalance is present in the \ac{OASIS3} benchmark, with 949 \ac{CN} and 288 \ac{AD} \ac{MRI} scans in the \ac{ID} data. 
Following the 70--15--15 split, 660 \ac{MRI} and 197 \ac{MRI} scans are available, respectively. 
To address this imbalance, 100 scans were randomly selected from the available \ac{CN} and \ac{AD} \ac{MRI} sessions per epoch.

\paragraph{Augmentation}
\begin{table}[htbp!]
\caption{The employed mean and standard deviation (SD) values for each dataset. For \ac{MIDOG} and \ac{PHAKIR}, the values correspond to the red, green, and blue channel. }
    \label{tab:normalization_values}
\centering
{\renewcommand\baselinestretch{1.0}\selectfont\resizebox{\linewidth}{!}{
 \begin{tabular}{l@{\hspace{13pt}}
    >{\centering\arraybackslash}c
    >{\centering\arraybackslash}c
    >{\centering\arraybackslash}c} 
\toprule              
     & \ac{MIDOG}     & \ac{PHAKIR}   & \ac{OASIS3}  \\
\midrule
Mean & 0.712\,/\,0.496\,/\,0.756 & 0.517\,/\,0.361\,/\,0.336 & \multirow{2}{*}{z-Normalization} \\
SD   & 0.167\,/\,0.167\,/\,0.110 & 0.166\,/\,0.143\,/\,0.137 & \\
\bottomrule
    \end{tabular}}\par}
\end{table}
To enhance classification performance on unseen data, each classifier was trained using additional data augmentations.

For the \ac{MIDOG} classifier, TrivialAugment Wide~\cite{muller2021trivialaugment} was chosen, as it encompasses a diverse range of augmentations. 
In contrast, for the \ac{PHAKIR} classifier, a custom augmentation pipeline provided the best results.
\begin{itemize}
    \item[] \textbf{Resize}: size=(360, 640)
    \item[] \textbf{RandomHorizontalFlip}: p=0.5
    \item[] \textbf{RandomPerspective}: distortion\textunderscore scale=0.2, p=0.5
    \item[] \textbf{ColorJitter}: brightness=0.2, contrast=0.2, saturation=0.1, hue=0.1
\end{itemize}
Similarly, the \ac{OASIS3} classifier was trained using a custom augmentation pipeline:
\begin{itemize}
    \item[] \textbf{RandomFlip}: axes='lr', p=0.5
    \item[] \textbf{RandomAffine}: scales=(0.9, 1.1), degrees=10, isotro-pic=True, default\textunderscore pad\textunderscore value='minimum', p=0.9
    \item[] \textbf{RandomMotion}: degrees=5, translation=5, p=0.2
    \item[] \textbf{RandomNoise}: std=(0, 0.1), p=0.9
    \item[] \textbf{RandomBlur}: std=(0, 0.2), p=0.9
    \item[] \textbf{RandomBiasField}: coefficients=(0.1, 0.3), p=0.8
    \item[] \textbf{RandomElasticDeformation}: max\textunderscore displacement=(5, 5, 5), p=0.1 
\end{itemize}
The final step in each data transformation pipeline was normalization, with the corresponding values provided in \cref{tab:normalization_values}.

\subsection{Results}
\label{sec:supplementary:experiments:results}
The \ac{AUROC} scores of the four top-performing methods on the most challenging datasets from the \ac{OASIS3} \ac{MIB}, \ac{ATLAS}, and \ac{BRATS} are presented in \cref{fig:top4_confidence_scores_oasis}. 

\begin{table}[bp!]
\caption{\ac{AUROC} performance averaged across all \ac{OOD} detection methods for each type. 
    The red percentages indicate the performance degradation compared to the best performing type Feature.}
    \label{tab:average_results}
\centering
{\renewcommand\baselinestretch{1.0}\selectfont\resizebox{\linewidth}{!}{
 \begin{tabular}{l@{\hspace{13pt}}
    >{\centering\arraybackslash}l
    >{\centering\arraybackslash}l
    >{\centering\arraybackslash}l} 
\toprule              
    Information source  & \ac{MIDOG}     & \ac{PHAKIR}   & \ac{OASIS3}  \\
\midrule

Feature & 66.08 & 70.68 & 92.08 \\
Combined & 57.18 \textcolor{red}{-13\,\%} & 50.71 \textcolor{red}{-28\,\%} & 69.86 \textcolor{red}{-24\,\%} \\
Classification & 55.81 \textcolor{red}{-16\,\%} & 49.45 \textcolor{red}{-30\,\%} & 52.13 \textcolor{red}{-43\,\%} \\

\bottomrule
    \end{tabular}}\par}
\end{table}
The substantial gap in discriminative power between classification- and hybrid-based methods, compared to feature-based methods, is shown in \cref{tab:average_results}.

To allow for an easier interpretation of the results, \cref{tab:method_descriptions_classification} -- \cref{tab:method_descriptions_hybrid} include descriptions for all evaluated \ac{OOD} methods.
\begin{table*}[bp!]
\caption{Description of each evaluated classification-based approach.}
    \label{tab:method_descriptions_classification}
\centering
{\renewcommand\baselinestretch{1.0}\selectfont\resizebox{\linewidth}{!}{
 \begin{tabular}{lp{0.85\linewidth}} 
\toprule              
    Method  & Description  \\
\midrule

\rowcolor{Classification_light} EBO~\cite{liu2020energy_EBO} & Motivated by energy-based learning \cite{lecun2006tutorial}, Liu \etal transform the final logits into a single scalar using the energy function $ E(\textbf{x};f)=- T\cdot \text{log}\sum_i^K e^{f_i(\textbf{x})/T}$. This scalar is then used as the confidence score for \ac{OOD} detection. \\
\rowcolor{Classification_light} Dropout~\cite{gal2016uncertainty_Dropout} & Based on the uncertainty estimation from Gal \etal, this method repeatedly sets entire channels from the penultimate feature layer to zero at random. The softmax probabilities of the resulting logits' mean are used as the confidence score. \\
\rowcolor{Classification_light} GEN~\cite{liu2023gen_GEN} & Liu \etal take the \emph{N} largest softmax probability and use the generalized entropy $G_\gamma(\textbf{p})=\sum_i p_{i}^{\gamma}(1-p_i)^\gamma$ as confidence score.\\
\rowcolor{Classification_light} KLM~\cite{hendrycks2019benchmark_KLM} & Hendrycks \etal compute the class-wise distribution of mean softmax probabilities. During inference the minimal Kullback-Leibler divergence between these mean distributions and the current sample distribution is used as confidence score.\\
\rowcolor{Classification_light} MLS~\cite{hendrycks2019scaling_MLS} & Instead of employing the maximum softmax probability, Hendrycks \etal use the maximum logit as confidence score.\\
\rowcolor{Classification_light} MSP~\cite{hendrycks2016baseline_MSP} & As one of the earliest baselines, Hendrycks \etal use the maximum softmax probability as confidence score.\\
\rowcolor{Classification_light} ODIN~\cite{liang2017enhancing_ODIN} & Liang \etal employ input perturbation and subsequent temperature scaling on the logits. Subsequently, the maximum softmax probability from these logits is used as confidence score. \\
\rowcolor{Classification_light} OpenMax~\cite{bendale2016towards_OpenMax} & Bendale and Boult first estimate Weibull distributions for all classes based on the top \emph{k} distances to mean logits. These distributions are used to rescale the logits. Subsequently an additional pseudo-logit is added, while the total activation level remains constant, which serves as \ac{OOD} class. The probability of this pseudo-class is used as confidence score. \\
\rowcolor{Classification_light} TempScale~\cite{guo2017calibration_TempScale} & Guo \etal learn a temperature scaling on the \ac{ID} dataset and use the temperature-scaled softmax probabilities as confidence score.\\

\bottomrule
    \end{tabular}}\par}
\end{table*}

\begin{table*}[bp!]
\caption{Description of each evaluated feature-based approach.}
    \label{tab:method_descriptions_feature}
\centering
{\renewcommand\baselinestretch{1.0}\selectfont\resizebox{\linewidth}{!}{
 \begin{tabular}{lp{0.85\linewidth}} 
\toprule              
    Method  & Description  \\
\midrule

\rowcolor{Feature_light} KNN~\cite{sun2022out_KNN} & Sun \etal compute the \emph{k}-th nearest neighbor of a sample inside the set of normalized activations from the penultimate layer. The distance to this neighbor is used as confidence score.\\
\rowcolor{Feature_light} MDS~\cite{lee2018simple_MDS} & Lee \etal uses the Mahalanobis distance between a sample's penultimate layer activations and class conditional Gaussian distributions derived from \ac{ID} data. \\
\rowcolor{Feature_light} MDSEns~\cite{lee2018simple_MDS} & Lee \etal extend MDS by aggregating these distances from all intermediate layers through weighted averaging. Additionally, input perturbation from ODIN is applied. \\
\rowcolor{Feature_light} Residual~\cite{wang2022vim_ViM_Residual} & Wang \etal project the activations from the penultimate layer to a low-variance subspace defined by the \emph{N} smallest eigenvalues of the empirical covariance matrix, estimated from \ac{ID} data and uses the norm of those activations as confidence score. \\
\rowcolor{Feature_light} RMDS~\cite{ren2021simple_RMDS} & Ren \etal extend MDS by introducing an additional Mahalanobis distance, computed between the penultimate layer activations and the Gaussian distribution estimated from the entire \ac{ID} dataset. The final confidence score is obtained by subtracting this new distance from the original MDS distances. \\
\rowcolor{Feature_light} SHE~\cite{zhang2022out_SHE} & Zhang \etal define confidence scores using the distance between a sample’s penultimate-layer activations and its class-conditional mean, computed exclusively from correctly classified samples. \\
\bottomrule
    \end{tabular}}\par}
\end{table*}

\begin{table*}[bp!]
\caption{Description of each evaluated hybrid-based approach.}
    \label{tab:method_descriptions_hybrid}
\centering
{\renewcommand\baselinestretch{1.0}\selectfont\resizebox{\linewidth}{!}{
 \begin{tabular}{lp{0.85\linewidth}} 
\toprule              
    Method  & Description  \\
\midrule
\rowcolor{Combined_light} ASH~\cite{djurisic2022extremely_ASH} & Djurisic \etal set the lowest \emph{p}th-percentile of activations in the penultimate layer to zero. The remaining activations are then processed in one of three ways: they are either left unchanged, replaced with a positive constant, or scaled by a ratio derived from the activations before and after pruning. Subsequently, these adjusted activations are used as input to the energy-score from EBO to yield the confidence score. We follow the implementation from OpenOOD~\cite{zhang2024openood} and use the variant with positive constants.\\
\rowcolor{Combined_light} DICE~\cite{sun2022dice_DICE} & Sun \etal calculate the class-wise contribution of weights in the final fully-connected layer based on the empirically estimated \ac{ID} mean of the \ac{ID} dataset. By preserving only the \emph{p}-th percentile of the most important weights, they calculate the final logits and use those as input to the energy-score from EBO to yield the confidence score. \\
\rowcolor{Combined_light} fDBD~\cite{liu2024_fdbd} & Liu \etal estimate the distance of the penultimate layers' features to class-decision boundaries and regularize this distance by the distance between the activation and the mean of activations from the \ac{ID} dataset. \\
\rowcolor{Combined_light} NNGuide~\cite{park2023nearest_NNGuide} & Park \etal use the average distance of a samples' activation from the penultimate layer to the \ac{ID} distribution of activations and use it to scale the energy-score from EBO to yield the confidence score. \\
\rowcolor{Combined_light} RankFeat~\cite{song2022rankfeat_RankFeat} & Song \etal propose to remove the rank-1 matrix from activations from the two last feature layers. These matrices are composed by the largest singular value established from a Singular Value Decomposition. The adjusted activations are forwared to yield new logits, which are then averaged and used as input to the energy-score from EBO to yield the confidence score. \\
\rowcolor{Combined_light} ReAct~\cite{sun2021react_ReAct} & Sun \etal calculate a threshold from the \emph{p}th-percentile of all \ac{ID} activations in the penultimate layer and use this threshold to set activations above this threshold to zero. Subsequently, logits resulting from these activations serve as input to the energy-score from EBO, yielding the final confidence score.\\
\rowcolor{Combined_light} Relation~\cite{kim2024neural_Relation} & Kim \etal estimate the relational structure on the feature-space of the penultimate layer based on the activations and corresponding class labels. This structure allows to identify similar feature embeddings with different label information. The confidence score is then calculated by evaluating their proposed similarity functions on a subset of the \ac{ID} data.\\
\rowcolor{Combined_light} SCALE~\cite{xu2023scaling_SCALE} & Motivated by ASH, Xu \etal omit the pruning step from ASH but keep the activation scaling based on the \emph{p}th-percentile of activations. These activations are then used as input to the energy-score from EBO to yield the confidence score.\\
\rowcolor{Combined_light} ViM~\cite{wang2022vim_ViM_Residual} & Wang \etal create an additional virtual-logit based on the subspace from Residual and calculate the energy-score over this and the original logits.  \\

\bottomrule
    \end{tabular}}\par}
\end{table*}

\begin{figure}[tbp!]
  \centering
    \includegraphics[width=1.0\linewidth]{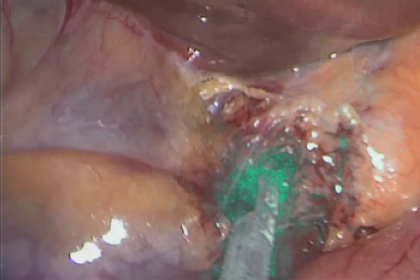}
    \begin{tikzpicture}[overlay, remember picture]        
        \draw[red, thick, ->, line width=2.5mm, >=stealth] (0.2, 4.0) -- (1.0, 3.5);
        \draw[red, thick, ->,line width=2.5mm, >=stealth] (1.07, 5.5) -- (1.87, 5.0);
    \end{tikzpicture}
    \caption{Image from the \ac{csID} Medium Smoke dataset. 
    Classification attribution is visualized in turquoise using Integrated Gradients (Sundarajan \etal~\cite{sundararajan2017axiomatic}), revealing the \ac{PHAKIR} classifier's tendency to base decisions on regions containing instruments. Arrows indicate an area with localized smoke.}
   \label{fig:attribution}
\end{figure}

\Cref{fig:attribution} shows that the \ac{PHAKIR} classifier predominantly bases its decisions on regions containing instruments. 
Thus, when smoke is located away from the instrument, it is likely that the feature embeddings are less influenced by the smoke.

\input{figures/feature_space}
In \cref{fig:phakir_tsne}, the t-SNE~\cite{van2008visualizing} visualization of the \ac{PHAKIR} classifier's feature space illustrates that \ac{OOD} samples from the \ac{EndoSeg18} dataset are primarily clustered near the Grasper class.

\input{figures/TPs_FPs}
\Cref{fig:TPs_FPs} presents success and failure cases for all \ac{OOD} settings.
For the \ac{MIDOG} and \ac{PHAKIR} benchmarks, these examples are derived from the two highest-ranked methods.
However, for \ac{OASIS3}, due to the lack of misclassifications in several \ac{OOD} categories for MDSEns and ViM, we selected the best performing methods that still exhibit failure cases in these scenarios: SHE and RMDS.

The remaining tables in this section (\cref{tab:midog_auroc} -- \cref{tab:imagenet_auprin_auprout}) provide detailed results for all \acp{MIB} and their corresponding datasets across all metrics as well as for the \ac{IN1k} benchmark from OpenOOD \cite{yang2022openood, yang2023full}. 
Entries in each table are sorted according to the overall results presented in \cref{tab:main_results}.

\begin{table*}[htbp]
    \caption{Results from the \acs{MIDOG} benchmark for the \acs{AUROC} and FPR@95 metrics.}
    \label{tab:midog_auroc}
\centering
{\renewcommand\baselinestretch{1.0}\selectfont\resizebox{\textwidth}{!}{
\begin{tabular}{l@{\hspace{13pt}}>{\centering\arraybackslash}m{0.95cm}>{\centering\arraybackslash}m{0.95cm}>{\centering\arraybackslash}m{0.95cm}@{\hspace{13pt}}>{\centering\arraybackslash}m{0.95cm}>{\centering\arraybackslash}m{0.95cm}>{\centering\arraybackslash}m{0.95cm}>{\centering\arraybackslash}m{0.95cm}>{\centering\arraybackslash}m{0.95cm}>{\centering\arraybackslash}m{0.95cm}>{\centering\arraybackslash}m{0.95cm}>{\centering\arraybackslash}m{0.95cm}@{\hspace{13pt}}
 >{\centering\arraybackslash}m{0.95cm}>{\centering\arraybackslash}m{0.95cm}>{\centering\arraybackslash}m{0.95cm}} 
        \toprule
        \multirow{2}{*}{} 
         & \multicolumn{3}{c}{\textbf{\acs{csID}}}
         & \multicolumn{8}{c}{\textbf{\acs{nOOD}}}
         & \multicolumn{3}{c}{\textbf{\acs{fOOD}}} \\
        
        & 1$_{\text{b}}$ & 1$_{\text{c}}$ & Avg 
        & 2 & 3 & 4 & 5 & 6$_{\text{a}}$ & 6$_{\text{b}}$ & 7 & Avg 
        & CCAgT & FNAC & Avg\\
    \midrule
\rowcolor{Gray_light} \textbf{\acs{AUROC}}\,\raisebox{0.3ex}{$\uparrow$} & & & & & & & & & & & & & &\\
\rowcolor{Feature_light} MDSEns~\cite{lee2018simple_MDS} & 98.56 & 99.82 & 99.19 & 99.56 & 99.71 & 99.68 & 71.50 & 98.95 & 99.71 & 73.78 & 91.84 & 100.00 & 100.00 & 100.00 \\
\rowcolor{Combined_light} ViM~\cite{wang2022vim_ViM_Residual} & 58.97 & 60.48 & 59.73 & 69.06 & 68.22 & 65.87 & 61.97 & 61.78 & 61.13 & 50.65 & 62.67 & 89.80 & 79.76 & 84.78 \\
\rowcolor{Feature_light} Residual~\cite{wang2022vim_ViM_Residual} & 57.87 & 62.65 & 60.26 & 72.56 & 72.15 & 67.57 & 64.48 & 65.21 & 63.56 & 54.89 & 65.78 & 94.80 & 89.91 & 92.35 \\
\rowcolor{Feature_light} MDS~\cite{lee2018simple_MDS} & 56.89 & 60.31 & 58.60 & 68.94 & 68.20 & 65.60 & 63.67 & 62.64 & 60.96 & 52.45 & 63.21 & 93.50 & 88.31 & 90.91 \\
\rowcolor{Feature_light} KNN~\cite{sun2022out_KNN} & 56.72 & 59.21 & 57.97 & 66.62 & 70.41 & 63.45 & 63.25 & 59.22 & 57.82 & 50.62 & 61.63 & 91.10 & 89.26 & 90.18 \\
\bottomrule\end{tabular}}}\end{table*}

\begin{table*}[htbp]
    \caption{Results from the \acs{MIDOG} benchmark for the \acs{AUPRIN} and \acs{AUPROUT} metrics. }
\centering
{\renewcommand\baselinestretch{1.0}\selectfont\resizebox{\textwidth}{!}{
 \begin{tabular}{l@{\hspace{13pt}}>{\centering\arraybackslash}m{0.95cm}>{\centering\arraybackslash}m{0.95cm}>{\centering\arraybackslash}m{0.95cm}@{\hspace{13pt}}>{\centering\arraybackslash}m{0.95cm}>{\centering\arraybackslash}m{0.95cm}>{\centering\arraybackslash}m{0.95cm}>{\centering\arraybackslash}m{0.95cm}>{\centering\arraybackslash}m{0.95cm}>{\centering\arraybackslash}m{0.95cm}>{\centering\arraybackslash}m{0.95cm}>{\centering\arraybackslash}m{0.95cm}@{\hspace{13pt}}
 >{\centering\arraybackslash}m{0.95cm}>{\centering\arraybackslash}m{0.95cm}>{\centering\arraybackslash}m{0.95cm}} 
        \toprule

        \multirow{2}{*}{} 
         & \multicolumn{3}{c}{\textbf{\acs{csID}}}
         & \multicolumn{8}{c}{\textbf{\acs{nOOD}}}
         & \multicolumn{3}{c}{\textbf{\acs{fOOD}}} \\
        
        & 1$_{\text{b}}$ & 1$_{\text{c}}$ & Avg 
        & 2 & 3 & 4 & 5 & 6$_{\text{a}}$ & 6$_{\text{b}}$ & 7 & Avg 
        & CCAgT & FNAC & Avg\\
    \midrule
\rowcolor{Gray_light} \textbf{\acs{AUPRIN}}\,\raisebox{0.3ex}{$\uparrow$} & & & & & & & & & & & & & &\\
\rowcolor{Feature_light} MDSEns~\cite{lee2018simple_MDS} & 95.64 & 99.28 & 97.46 & 98.85 & 98.83 & 98.86 & 13.30 & 97.90 & 99.55 & 13.16 & 74.35 & 100.00 & 99.99 & 100.00 \\
\rowcolor{Combined_light} ViM~\cite{wang2022vim_ViM_Residual} & 10.20 & 12.75 & 11.48 & 16.55 & 5.27 & 9.69 & 10.49 & 7.24 & 30.35 & 7.58 & 12.45 & 44.07 & 57.24 & 50.65 \\
\rowcolor{Feature_light} Residual~\cite{wang2022vim_ViM_Residual} & 10.35 & 13.69 & 12.02 & 19.55 & 6.85 & 12.35 & 10.88 & 8.92 & 34.45 & 8.00 & 14.43 & 55.84 & 72.06 & 63.95 \\
\rowcolor{Feature_light} MDS~\cite{lee2018simple_MDS} & 10.46 & 13.03 & 11.74 & 19.06 & 7.30 & 12.41 & 11.35 & 8.78 & 32.40 & 7.32 & 14.09 & 51.93 & 67.59 & 59.76 \\
\rowcolor{Feature_light} KNN~\cite{sun2022out_KNN} & 11.81 & 11.15 & 11.48 & 15.55 & 11.42 & 9.60 & 10.33 & 6.15 & 31.42 & 6.33 & 12.97 & 28.37 & 65.82 & 47.10 \\
\bottomrule
    \end{tabular}}\par}
\end{table*}

\begin{table*}[htbp]
    \caption{Results from the \acs{PHAKIR} benchmark for the \acs{AUROC} and FPR@95 metrics. 
    M. Smoke and H. Smoke stands for Medium and Heavy Smoke.
    CAT. stands for CATARACTS.}
\centering
{\renewcommand\baselinestretch{1.0}\selectfont\resizebox{\textwidth}{!}{
\begin{tabular}{l@{\hspace{13pt}}>{\centering\arraybackslash}m{1.51cm}>{\centering\arraybackslash}m{1.51cm}>{\centering\arraybackslash}m{1.51cm}@{\hspace{13pt}}>{\centering\arraybackslash}m{1.51cm}>{\centering\arraybackslash}m{1.51cm}>{\centering\arraybackslash}m{1.51cm}>{\centering\arraybackslash}m{1.51cm}@{\hspace{13pt}}>{\centering\arraybackslash}m{1.51cm}>{\centering\arraybackslash}m{1.51cm}>{\centering\arraybackslash}m{1.51cm}} 
        \toprule

        \multirow{2}{*}{} 
         & \multicolumn{3}{c}{\textbf{\acs{csID}}}
         & \multicolumn{4}{c}{\textbf{\acs{nOOD}}}
         & \multicolumn{3}{c}{\textbf{\acs{fOOD}}} \\
        
        & {M. Smoke} & {H. Smoke} & Avg 
        & \acs{cholec} & \acs{EndoSeg15} & \acs{EndoSeg18} & Avg 
        & Kvasir & CAT. & Avg\\
    \midrule
\rowcolor{Gray_light} \textbf{\acs{AUROC}}\,\raisebox{0.3ex}{$\uparrow$} & & & & & & & & & & \\
\rowcolor{Feature_light} MDSEns~\cite{lee2018simple_MDS} & 45.67 & 84.43 & 65.05 & 96.65 & 94.81 & 99.87 & 97.11 & 99.98 & 97.02 & 98.50 \\
\rowcolor{Combined_light} ViM~\cite{wang2022vim_ViM_Residual} & 63.64 & 81.14 & 72.39 & 68.44 & 83.25 & 91.73 & 81.14 & 50.04 & 60.64 & 55.34 \\
\rowcolor{Feature_light} Residual~\cite{wang2022vim_ViM_Residual} & 39.11 & 75.13 & 57.12 & 59.53 & 75.98 & 95.46 & 76.99 & 48.33 & 66.30 & 57.31 \\
\rowcolor{Feature_light} MDS~\cite{lee2018simple_MDS} & 38.25 & 73.84 & 56.04 & 58.27 & 76.37 & 94.81 & 76.48 & 40.20 & 62.75 & 51.47 \\
\rowcolor{Feature_light} KNN~\cite{sun2022out_KNN} & 24.01 & 44.09 & 34.05 & 64.17 & 61.22 & 40.92 & 55.44 & 31.22 & 44.28 & 37.75 \\
\bottomrule
    \end{tabular}}\par}
\end{table*}

\begin{table*}[htbp]
    \caption{Results from the \acs{PHAKIR} benchmark for the \acs{AUPRIN} and \acs{AUPROUT} metrics. M. Smoke and H. Smoke stands for Medium and Heavy Smoke.
    CAT. stands for CATARACTS.}
    \label{tab:phakir_auroc_fpr}
\centering
{\renewcommand\baselinestretch{1.0}\selectfont\resizebox{\textwidth}{!}{
\begin{tabular}{l@{\hspace{13pt}}>{\centering\arraybackslash}m{1.51cm}>{\centering\arraybackslash}m{1.51cm}>{\centering\arraybackslash}m{1.51cm}@{\hspace{13pt}}>{\centering\arraybackslash}m{1.51cm}>{\centering\arraybackslash}m{1.51cm}>{\centering\arraybackslash}m{1.51cm}>{\centering\arraybackslash}m{1.51cm}@{\hspace{13pt}}>{\centering\arraybackslash}m{1.51cm}>{\centering\arraybackslash}m{1.51cm}>{\centering\arraybackslash}m{1.51cm}} 
        \toprule

        \multirow{2}{*}{} 
         & \multicolumn{3}{c}{\textbf{\acs{csID}}}
         & \multicolumn{4}{c}{\textbf{\acs{nOOD}}}
         & \multicolumn{3}{c}{\textbf{\acs{fOOD}}} \\
        
        & {M. Smoke} & {H. Smoke} & Avg 
        & \acs{cholec} & \acs{EndoSeg15} & \acs{EndoSeg18} & Avg 
        & Kvasir & CAT. & Avg\\
    \midrule
\rowcolor{Gray_light} \textbf{\acs{AUPRIN}}\,\raisebox{0.3ex}{$\uparrow$} & & & & & & & & & & \\
\rowcolor{Feature_light} MDSEns~\cite{lee2018simple_MDS} & 29.60 & 69.21 & 49.41 & 73.26 & 98.59 & 99.80 & 90.55 & 99.96 & 87.55 & 93.75 \\
\rowcolor{Combined_light} ViM~\cite{wang2022vim_ViM_Residual} & 40.23 & 61.00 & 50.61 & 2.46 & 94.66 & 89.08 & 62.07 & 31.28 & 0.82 & 16.05 \\
\rowcolor{Feature_light} Residual~\cite{wang2022vim_ViM_Residual} & 27.14 & 52.52 & 39.83 & 1.16 & 92.18 & 94.58 & 62.64 & 32.41 & 1.11 & 16.76 \\
\rowcolor{Feature_light} MDS~\cite{lee2018simple_MDS} & 26.85 & 51.74 & 39.29 & 1.10 & 92.31 & 93.55 & 62.32 & 27.96 & 1.02 & 14.49 \\
\rowcolor{Feature_light} KNN~\cite{sun2022out_KNN} & 22.77 & 27.39 & 25.08 & 1.41 & 84.54 & 32.37 & 39.44 & 26.53 & 0.33 & 13.43 \\
\bottomrule
    \end{tabular}}\par}
\end{table*}

\begin{table*}[htbp]
    \caption{Results from the \acs{OASIS3} benchmark for the \acs{AUROC} and FPR@95 metrics. }
\centering
{\renewcommand\baselinestretch{1.0}\selectfont\resizebox{\textwidth}{!}{
\begin{tabular}{l@{\hspace{13pt}}>{\centering\arraybackslash}m{1.51cm}>{\centering\arraybackslash}m{1.51cm}>{\centering\arraybackslash}m{1.51cm}@{\hspace{13pt}}>{\centering\arraybackslash}m{1.51cm}>{\centering\arraybackslash}m{1.51cm}>{\centering\arraybackslash}m{1.51cm}>{\centering\arraybackslash}m{1.51cm}@{\hspace{13pt}}>{\centering\arraybackslash}m{1.51cm}>{\centering\arraybackslash}m{1.51cm}>{\centering\arraybackslash}m{1.51cm}} 
        \toprule

        \multirow{2}{*}{} 
         & \multicolumn{3}{c}{\textbf{\acs{csID}}}
         & \multicolumn{4}{c}{\textbf{\acs{nOOD}}}
         & \multicolumn{3}{c}{\textbf{\acs{fOOD}}} \\
        
        & Modality & Scanner & Avg 
        & \acs{ATLAS} & \acs{BRATS} & CT & Avg 
        & \acs{MSD-H} & \acs{CHAOS} & Avg \\
    \midrule
\rowcolor{Gray_light} \textbf{\acs{AUROC}}\,\raisebox{0.3ex}{$\uparrow$} & & & & & & & & & & \\
\rowcolor{Feature_light} MDSEns~\cite{lee2018simple_MDS} & 100.00 & 100.00 & 100.00 & 98.47 & 99.98 & 99.94 & 99.46 & 100.00 & 100.00 & 100.00 \\
\rowcolor{Combined_light} ViM~\cite{wang2022vim_ViM_Residual} & 100.00 & 97.63 & 98.82 & 98.19 & 97.01 & 100.00 & 98.40 & 100.00 & 100.00 & 100.00 \\
\rowcolor{Feature_light} Residual~\cite{wang2022vim_ViM_Residual} & 100.00 & 93.94 & 96.97 & 95.38 & 94.72 & 100.00 & 96.70 & 100.00 & 100.00 & 100.00 \\
\rowcolor{Feature_light} MDS~\cite{lee2018simple_MDS} & 100.00 & 92.62 & 96.31 & 94.65 & 93.80 & 100.00 & 96.15 & 100.00 & 100.00 & 100.00 \\
\rowcolor{Feature_light} KNN~\cite{sun2022out_KNN} & 100.00 & 97.56 & 98.78 & 96.25 & 96.71 & 100.00 & 97.66 & 100.00 & 100.00 & 100.00 \\
\bottomrule
    \end{tabular}}\par}
\end{table*}

\begin{table*}[htbp]
    \caption{Results from the \acs{OASIS3} benchmark for the \acs{AUPRIN} and \acs{AUPROUT} metrics. }
    \label{tab:oasis_aupr}
\centering
{\renewcommand\baselinestretch{1.0}\selectfont\resizebox{\textwidth}{!}{
\begin{tabular}{l@{\hspace{13pt}}>{\centering\arraybackslash}m{1.51cm}>{\centering\arraybackslash}m{1.51cm}>{\centering\arraybackslash}m{1.51cm}@{\hspace{13pt}}>{\centering\arraybackslash}m{1.51cm}>{\centering\arraybackslash}m{1.51cm}>{\centering\arraybackslash}m{1.51cm}>{\centering\arraybackslash}m{1.51cm}@{\hspace{13pt}}>{\centering\arraybackslash}m{1.51cm}>{\centering\arraybackslash}m{1.51cm}>{\centering\arraybackslash}m{1.51cm}} 
        \toprule
        \multirow{2}{*}{} 
         & \multicolumn{3}{c}{\textbf{\acs{csID}}}
         & \multicolumn{4}{c}{\textbf{\acs{nOOD}}}
         & \multicolumn{3}{c}{\textbf{\acs{fOOD}}} \\
        & Modality & Scanner & Avg 
        & \acs{ATLAS} & \acs{BRATS} & CT & Avg 
        & \acs{MSD-H} & \acs{CHAOS} & Avg \\
    \midrule
\rowcolor{Gray_light} \textbf{\acs{AUPRIN}}\,\raisebox{0.3ex}{$\uparrow$} & & & & & & & & & & \\
\rowcolor{Feature_light} MDSEns~\cite{lee2018simple_MDS} & 100.00 & 100.00 & 100.00 & 85.77 & 99.89 & 99.22 & 94.96 & 100.00 & 100.00 & 100.00 \\
\rowcolor{Combined_light} ViM~\cite{wang2022vim_ViM_Residual} & 100.00 & 98.95 & 99.47 & 95.50 & 89.45 & 100.00 & 94.99 & 100.00 & 100.00 & 100.00 \\
\rowcolor{Feature_light} Residual~\cite{wang2022vim_ViM_Residual} & 100.00 & 97.54 & 98.77 & 93.47 & 87.17 & 100.00 & 93.55 & 100.00 & 100.00 & 100.00 \\
\rowcolor{Feature_light} MDS~\cite{lee2018simple_MDS} & 99.99 & 96.98 & 98.49 & 92.51 & 85.57 & 100.00 & 92.69 & 100.00 & 100.00 & 100.00 \\
\rowcolor{Feature_light} KNN~\cite{sun2022out_KNN} & 100.00 & 98.93 & 99.46 & 92.93 & 88.62 & 100.00 & 93.85 & 100.00 & 100.00 & 100.00 \\
\bottomrule
    \end{tabular}}\par}
\end{table*}

\begin{table*}[htbp]
    \caption{Results from the \acs{IN1k} benchmark for the \acs{AUROC} and FPR@95 metrics.}
    \label{tab:imagenet_auroc_fpr95}
\centering
{\renewcommand\baselinestretch{1.0}\selectfont\resizebox{\textwidth}{!}{
 \begin{tabular}{l@{\hspace{13pt}}>{\centering\arraybackslash}m{1.45cm}>{\centering\arraybackslash}m{1.45cm}>{\centering\arraybackslash}m{1.45cm}>{\centering\arraybackslash}m{1.45cm}@{\hspace{13pt}}
 >{\centering\arraybackslash}m{1.45cm}>{\centering\arraybackslash}m{1.45cm}>{\centering\arraybackslash}m{1.45cm}@{\hspace{13pt}}
 >{\centering\arraybackslash}m{1.45cm}>{\centering\arraybackslash}m{1.45cm}>{\centering\arraybackslash}m{1.45cm}>{\centering\arraybackslash}m{1.45cm}} 
        \toprule

        \multirow{2}{*}{} 
         & \multicolumn{4}{c}{\textbf{\acs{csID}}}
         & \multicolumn{3}{c}{\textbf{\acs{nOOD}}}
         & \multicolumn{4}{c}{\textbf{\acs{fOOD}}} \\
        
        & IN-V2 \cite{recht2019imagenet} & IN-C \,~\cite{hendrycks2018benchmarking} & IN-R \,~\cite{hendrycks2021many} & Avg 
        & SSB-hard \cite{vaze2022openset} & NINCO \cite{bitterwolf2023inorout} & Avg 
        & iNaturalist \cite{van2018inaturalist} & Textures \cite{cimpoi2014describing} & OpenImage-O \cite{wang2022vim_ViM_Residual} & Avg\\
    \midrule
\rowcolor{Gray_light} \textbf{\acs{AUROC}}\,\raisebox{0.3ex}{$\uparrow$} & & & & & & & & & & &\\
\rowcolor{Feature_light} MDSEns~\cite{lee2018simple_MDS} & 51.15 & 76.74 & 74.87 & 67.58 & 48.30 & 60.66 & 54.48 & 56.42 & 93.30 & 73.82 & 74.51 \\
\rowcolor{Combined_light} ViM~\cite{wang2022vim_ViM_Residual} & 57.34 & 83.77 & 87.95 & 76.35 & 65.53 & 78.63 & 72.08 & 89.56 & 97.97 & 90.50 & 92.68 \\
\rowcolor{Feature_light} Residual~\cite{wang2022vim_ViM_Residual} & 49.83 & 67.78 & 65.52 & 61.04 & 42.14 & 54.59 & 48.37 & 52.13 & 87.81 & 61.02 & 66.99 \\
\rowcolor{Feature_light} MDS~\cite{lee2018simple_MDS} & 51.72 & 70.80 & 69.78 & 64.10 & 48.50 & 62.38 & 55.44 & 63.67 & 89.80 & 69.27 & 74.25 \\
\rowcolor{Feature_light} KNN~\cite{sun2022out_KNN} & 56.44 & 83.94 & 87.64 & 76.01 & 62.57 & 79.64 & 71.10 & 86.41 & 97.09 & 87.04 & 90.18 \\
\bottomrule
    \end{tabular}}\par}
\end{table*}

\begin{table*}[htbp]
    \caption{Results from the \acs{IN1k} benchmark for the \acs{AUPRIN} and \acs{AUPROUT} metrics. }
    \label{tab:imagenet_auprin_auprout}
\centering
{\renewcommand\baselinestretch{1.0}\selectfont\resizebox{\textwidth}{!}{
 \begin{tabular}{l@{\hspace{13pt}}>{\centering\arraybackslash}m{1.45cm}>{\centering\arraybackslash}m{1.45cm}>{\centering\arraybackslash}m{1.45cm}>{\centering\arraybackslash}m{1.45cm}@{\hspace{13pt}}
 >{\centering\arraybackslash}m{1.45cm}>{\centering\arraybackslash}m{1.45cm}>{\centering\arraybackslash}m{1.45cm}@{\hspace{13pt}}
 >{\centering\arraybackslash}m{1.45cm}>{\centering\arraybackslash}m{1.45cm}>{\centering\arraybackslash}m{1.45cm}>{\centering\arraybackslash}m{1.45cm}} 
        \toprule

        \multirow{2}{*}{} 
         & \multicolumn{4}{c}{\textbf{\acs{csID}}}
         & \multicolumn{3}{c}{\textbf{\acs{nOOD}}}
         & \multicolumn{4}{c}{\textbf{\acs{fOOD}}} \\
        
        & IN-V2 \cite{recht2019imagenet} & IN-C \,~\cite{hendrycks2018benchmarking} & IN-R \,~\cite{hendrycks2021many} & Avg 
        & SSB-hard \cite{vaze2022openset} & NINCO \cite{bitterwolf2023inorout} & Avg 
        & iNaturalist \cite{van2018inaturalist} & Textures \cite{cimpoi2014describing} & OpenImage-O \cite{wang2022vim_ViM_Residual} & Avg\\
    \midrule
\rowcolor{Gray_light} \textbf{\acs{AUPRIN}}\,\raisebox{0.3ex}{$\uparrow$} & & & & & & & & & & &\\
\rowcolor{Feature_light} MDSEns~\cite{lee2018simple_MDS} & 82.96 & 91.56 & 82.45 & 85.66 & 48.23 & 92.32 & 70.28 & 87.13 & 98.47 & 88.79 & 91.46 \\
\rowcolor{Combined_light} ViM~\cite{wang2022vim_ViM_Residual} & 85.14 & 95.03 & 91.50 & 90.56 & 65.67 & 96.49 & 81.08 & 97.59 & 99.74 & 96.58 & 97.97 \\
\rowcolor{Feature_light} Residual~\cite{wang2022vim_ViM_Residual} & 82.20 & 90.00 & 75.91 & 82.70 & 42.63 & 90.66 & 66.65 & 84.35 & 98.17 & 81.52 & 88.01 \\
\rowcolor{Feature_light} MDS~\cite{lee2018simple_MDS} & 83.11 & 91.37 & 79.81 & 84.76 & 49.37 & 93.24 & 71.30 & 89.91 & 98.60 & 87.35 & 91.95 \\
\rowcolor{Feature_light} KNN~\cite{sun2022out_KNN} & 84.66 & 95.01 & 90.80 & 90.16 & 62.33 & 96.74 & 79.53 & 96.66 & 99.61 & 95.04 & 97.10 \\
\bottomrule
    \end{tabular}}\par}
\end{table*}

\subsection{Employed Hyperparameters}
\label{sec:supplementary:experiments:hp}
\begin{table*}[htbp]
\caption{Overview of the hyperparameter ranges for each \ac{OOD} detection method with tunable hyperparameters employed in this work.}
    \label{tab:hyperparameters_range}
\centering
{\renewcommand\baselinestretch{1.0}\selectfont\resizebox{\textwidth}{!}{
\begin{tabular}{lll@{\hspace{20pt}}ll}
\toprule
 Method & Parameter 1      & Parameter Range & Parameter 2 & Parameter Range \\ 
\midrule
MDSEns           &  noise       & [0, 0.0025, 0.0014, 0.005, 0.01, 0.02, 0.04, 0.08]    & -- & \\
ViM              &  dim         & [1, 16, 32, 64, 128, 256]                             & -- & \\
Residual         &  dim         & [32, 64, 128, 256, 512, 1024]                         & -- & \\
KNN              &  k           & [1, 2, 5, 10, 25, 50, 100, 200, 500, 750, 1000]       & -- & \\
SHE              &  metric      & [inner product, euclidean, cosine]     & -- & \\ 
Relation         &  pow         & [1, 2, 4, 6, 8]                                       & -- & \\
fDBD             &  normalized  & [False,True]                                          & -- & \\
\bottomrule
    \end{tabular}}\par}
\end{table*}
\begin{table*}[htbp]
\caption{Overview of the automatically selected hyperparameters for the three \acp{MIB} and \ac{IN1k}.}
    \label{tab:hyperparameters}
\centering
{\renewcommand\baselinestretch{1.0}\selectfont\resizebox{\textwidth}{!}{
\begin{tabular}{lll@{\hspace{20pt}}ll@{\hspace{20pt}}ll@{\hspace{20pt}}ll}
\toprule
\multirow{2}{*}{} & \multicolumn{2}{c}{\ac{MIDOG}} & \multicolumn{2}{c}{\ac{PHAKIR}} & \multicolumn{2}{c}{\ac{OASIS3}} & \multicolumn{2}{c}{\ac{IN1k}} \\
  Method                & Parameter 1 & Parameter 2 & Parameter 1 & Parameter 2 & Parameter 1 & Parameter 2 & Parameter 1 & Parameter 2 \\ 
\midrule
MDSEns           &  0.0                  & --                 & 0.0014                 & --                 & 0.0 & -- & 0.0 & --          \\
ViM              &  256                  & --                 & 256                 & --                 & 256 & -- & 256 & --                \\
Residual         &  128                   & --                 & 128                 & --                 & 64 & -- & 1024 & --               \\
KNN              &  5                 & --                 & 50                 & --                 & 5 & -- & 200 & --                \\
SHE              &  cosine                 & --                  & cosine & --                  & cosine & -- & cosine & --                \\ 
Relation         &  8               & --                & 8                 & --                 & 8 & -- & 1 & --                \\
\bottomrule
    \end{tabular}}\par}
\end{table*}
Most \ac{OOD} detection methods rely on one or more hyperparameters to optimize their performance by using the \ac{OOD} validation set. 
The search space for each method’s parameters is outlined in \cref{tab:hyperparameters_range}. 
Methods not included in this table either do not have tunable hyperparameters or feature parameters that are not easily adjustable. 
The parameter ranges are based on the OpenOOD framework~\cite{yang2022openood, zhang2024openood}, with minor adjustments. 
\Cref{tab:hyperparameters} provides a summary of all automatically selected hyperparameters for the \ac{OOD} detection methods.

\clearpage
\setcounter{page}{1}
\raggedbottom      
\setlength{\parskip}{0.4em}
\setlength{\parindent}{0pt}

\twocolumn[{%
\vspace*{1cm}
\begin{center}
    {\Large \bfseries Corrigendum to}\\[0.6em]
    {\large ``OpenMIBOOD: Open Medical Imaging Benchmarks for Out-of-Distribution Detection''}
\end{center}
\vspace{0.6cm}
}]
\label{sec:corrigendum}

\subsection*{Abstract}
During continued development of OpenMIBOOD and follow-up validation experiments, we identified implementation issues that affected the reported results of two OOD detection methods: NNGuide and ViM. 
These issues relate to the handling of internal model states after hyperparameter optimization. 
This corrigendum provides corrected results and clarifies the scope of the corrections. 
The main findings and conclusions of the original paper remain unchanged.

\subsection*{Background}
OpenMIBOOD was built upon the OpenOOD framework, which we extended and modified for medical imaging benchmarks.
Throughout development, we continuously validated and improved the framework implementation to ensure consistent results across all benchmarks.
However, after release, further review revealed additional issues affecting the reported results of NNGuide and ViM.

\subsection*{Description of the Issue}
The root cause was an inconsistency between the internal states used during hyperparameter search and those used during evaluation.
Specifically, some methods maintain internal statistics and variables that were not always recomputed after hyperparameter optimization, leading to incorrect evaluations in certain cases.

\subsection*{Affected Results}
\paragraph{ViM} 
A bug was identified during the publication process. 
All experiments were re-run, and the corrected results were incorporated for OASIS-3 and ImageNet-1k. 
However, due to an oversight, the previous (uncorrected) results for MIDOG and PhaKIR were inadvertently retained.

\paragraph{NNGuide}
A bug was found after publication, invalidating previously reported results for this approach.

We have re-executed all affected experiments and provide corrected tables and figures in this corrigendum. 
Results for all other methods and datasets remain unchanged.

To be precise, this corrigendum includes new versions of \cref{tab:main_results}, \cref{fig:top4_confidence_scores}, \cref{fig:benchmarks_vs_imagenet}, \cref{tab:average_results}, \cref{tab:midog_auroc}--\cref{tab:imagenet_auprin_auprout}, and \cref{tab:hyperparameters}.
We provide the full tables to reflect the updated overall ranking of ViM.

\subsection*{Impact on Conclusions}
While ViM previously ranked second and it is now on fourth rank, the key discussion on the top 4 ranked methods stays intact.
Residual and MDS each improve by one rank. 
Our key observations regarding the difficulty of OOD detection in medical imaging and the general performance trends across datasets remain valid.
The performance changes of NNGuide are negligible.
\input{figures/top4_confidence_scores_vim_corrected}

\subsection*{Acknowledgment}
We take full responsibility for the correctness of the reported results and appreciate the community’s engagement in verifying and improving reproducibility.
We hope this update reinforces confidence in the reliability and transparency of the OpenMIBOOD benchmarks.

\newpage

\input{figures/mib_vs_in_vim_nnguide_corrected}
\begin{table*}[htbp]
    \caption{
    Results of all evaluated \acs{OOD} detection methods, ranked by their average performance (\acs{AUROC}) across all \acs{MIB} groups for the \textbf{\acs{nOOD}} category.
    Additionally, the table includes the averaged \acs{nOOD} performance based on the harmonic mean of \acl{AUPRIN} and \acl{AUPROUT}, as well as the averaged \acs{FPR}@95 metrics. The F1-Score indicates the performance on the \acs{ID} test set.
    Rows are color-coded to indicate the source of information used for \acs{OOD} detection: blue for probabilities or logits, orange for feature space, and green for a combination of both.
    *: The results of the MDSEns method are potentially misleading, as discussed in the main text (\cref{par:top-performing}).}
    \label{tab:main_results_vim_nnguide_corrected}
\centering
{\renewcommand\baselinestretch{1.0}\selectfont\resizebox{\textwidth}{!}{
\begin{tabular}{l@{\hspace{13pt}}
    >{\centering\arraybackslash\color{dark_gr}}m{1.48cm}>{\centering\arraybackslash}m{1.48cm}>{\centering\arraybackslash\color{dark_gr}}m{1.48cm}@{\hspace{13pt}}
     >{\centering\arraybackslash\color{dark_gr}}m{1.48cm}>{\centering\arraybackslash}m{1.48cm}>{\centering\arraybackslash\color{dark_gr}}m{1.48cm}@{\hspace{13pt}}
     >{\centering\arraybackslash\color{dark_gr}}m{1.48cm}>{\centering\arraybackslash}m{1.48cm}>{\centering\arraybackslash\color{dark_gr}}m{1.48cm}@{\hspace{13pt}}
    c>{\color{dark_gr}}c>{\color{dark_gr}}c} 
        \toprule
        \multirow{2}{*}{} 
         & \multicolumn{3}{c}{\textbf{\acs{MIDOG}} @ F1-Score: \num{81.88}} 
         & \multicolumn{3}{c}{\textbf{\acs{PHAKIR}} @ F1-Score: \num{80.08}} 
         & \multicolumn{3}{c}{\textbf{\acs{OASIS3}} @ F1-Score: \num{73.65}} 
         & \multicolumn{3}{c}{\textbf{Averages for \acs{nOOD}}} \\
        
        & \acs{csID} & \tikzmark{midog_topLeft}\textcolor{black}{\acs{nOOD}} & \acs{fOOD} & \acs{csID} & \tikzmark{phakir_topLeft}\textcolor{black}{\acs{nOOD}} & \acs{fOOD} & \acs{csID} & \tikzmark{oasis_topLeft}\textcolor{black}{\acs{nOOD}} & \acs{fOOD} & \tikzmark{avg_topLeft}$\overline{\text{\acs{AUROC}}}$\,\raisebox{0.3ex}{$\uparrow$} & 
        $\overline{\text{\acs{AUPR}}}$\,\raisebox{0.3ex}{$\uparrow$} & $\overline{\text{\acs{FPR}@95}}$\,\raisebox{0.3ex}{$\downarrow$}\\
    \midrule
\rowcolor{Feature_light} MDSEns*~\cite{lee2018simple_MDS}	& 99.19 & 91.84 & 100.00	& 65.05 & 97.11 & 98.50	& 100.00 & 99.46 & 100.00 & 96.14 & 91.86 & 11.97 \\
\rowcolor{Feature_light} Residual~\cite{wang2022vim_ViM_Residual}	& 60.26 & 65.78 & 92.35	& 57.12 & 76.99 & 57.31	& 96.97 & 96.70 & 100.00 & 79.82 & 70.18 & 49.94 \\
\bottomrule
    \end{tabular}}\par}
\end{table*}
\begin{table}[bp!]
\caption{\ac{AUROC} performance averaged across all \ac{OOD} detection methods for each type. 
    The red percentages indicate the performance degradation compared to the best performing type Feature.}
    \label{tab:average_results_vim_nnguide_corrected}
\centering
{\renewcommand\baselinestretch{1.0}\selectfont\resizebox{\linewidth}{!}{
 \begin{tabular}{l@{\hspace{13pt}}
    >{\centering\arraybackslash}l
    >{\centering\arraybackslash}l
    >{\centering\arraybackslash}l} 
\toprule              
    Information source  & \ac{MIDOG}     & \ac{PHAKIR}   & \ac{OASIS3}  \\
\midrule

Feature & 66.08 & 70.68 & 92.08 \\
Combined & 57.34 \textcolor{red}{-13\,\%} & 48.86 \textcolor{red}{-31\,\%} & 69.62 \textcolor{red}{-24\,\%} \\
Classification & 55.81 \textcolor{red}{-16\,\%} & 49.45 \textcolor{red}{-30\,\%} & 52.13 \textcolor{red}{-43\,\%} \\

\bottomrule
    \end{tabular}}\par}
\end{table}

\begin{table*}[htbp]
    \caption{Results from the \acs{MIDOG} benchmark for the \acs{AUROC} and FPR@95 metrics.}
    \label{tab:midog_auroc_vim_nnguide_corrected}
\centering
{\renewcommand\baselinestretch{1.0}\selectfont\resizebox{\textwidth}{!}{
 \begin{tabular}{l@{\hspace{13pt}}>{\centering\arraybackslash}m{0.95cm}>{\centering\arraybackslash}m{0.95cm}>{\centering\arraybackslash}m{0.95cm}@{\hspace{13pt}}>{\centering\arraybackslash}m{0.95cm}>{\centering\arraybackslash}m{0.95cm}>{\centering\arraybackslash}m{0.95cm}>{\centering\arraybackslash}m{0.95cm}>{\centering\arraybackslash}m{0.95cm}>{\centering\arraybackslash}m{0.95cm}>{\centering\arraybackslash}m{0.95cm}>{\centering\arraybackslash}m{0.95cm}@{\hspace{13pt}}
 >{\centering\arraybackslash}m{0.95cm}>{\centering\arraybackslash}m{0.95cm}>{\centering\arraybackslash}m{0.95cm}} 
        \toprule
        \multirow{2}{*}{} 
         & \multicolumn{3}{c}{\textbf{\acs{csID}}}
         & \multicolumn{8}{c}{\textbf{\acs{nOOD}}}
         & \multicolumn{3}{c}{\textbf{\acs{fOOD}}} \\
        
        & 1$_{\text{b}}$ & 1$_{\text{c}}$ & Avg 
        & 2 & 3 & 4 & 5 & 6$_{\text{a}}$ & 6$_{\text{b}}$ & 7 & Avg 
        & CCAgT & FNAC & Avg\\
    \midrule
\rowcolor{Gray_light} \textbf{\acs{AUROC}}\,\raisebox{0.3ex}{$\uparrow$} & & & & & & & & & & & & & &\\
\rowcolor{Feature_light} MDSEns~\cite{lee2018simple_MDS} & 98.56 & 99.82 & 99.19 & 99.56 & 99.71 & 99.68 & 71.50 & 98.95 & 99.71 & 73.78 & 91.84 & 100.00 & 100.00 & 100.00 \\
\rowcolor{Feature_light} Residual~\cite{wang2022vim_ViM_Residual} & 57.87 & 62.65 & 60.26 & 72.56 & 72.15 & 67.57 & 64.48 & 65.21 & 63.56 & 54.89 & 65.78 & 94.80 & 89.91 & 92.35 \\
\rowcolor{Feature_light} MDS~\cite{lee2018simple_MDS} & 56.89 & 60.31 & 58.60 & 68.94 & 68.20 & 65.60 & 63.67 & 62.64 & 60.96 & 52.45 & 63.21 & 93.50 & 88.31 & 90.91 \\
\rowcolor{Combined_light} ViM~\cite{wang2022vim_ViM_Residual} & 56.75 & 61.68 & 59.22 & 69.92 & 70.57 & 65.54 & 64.70 & 63.13 & 61.76 & 54.86 & 64.35 & 94.53 & 91.69 & 93.11 \\
\rowcolor{Feature_light} KNN~\cite{sun2022out_KNN} & 56.72 & 59.21 & 57.97 & 66.62 & 70.41 & 63.45 & 63.25 & 59.22 & 57.82 & 50.62 & 61.63 & 91.10 & 89.26 & 90.18 \\
\bottomrule\end{tabular}}}\end{table*}

\begin{table*}[htbp]
    \caption{Results from the \acs{MIDOG} benchmark for the \acs{AUPRIN} and \acs{AUPROUT} metrics. }
\centering
{\renewcommand\baselinestretch{1.0}\selectfont\resizebox{\textwidth}{!}{
 \begin{tabular}{l@{\hspace{13pt}}>{\centering\arraybackslash}m{0.95cm}>{\centering\arraybackslash}m{0.95cm}>{\centering\arraybackslash}m{0.95cm}@{\hspace{13pt}}>{\centering\arraybackslash}m{0.95cm}>{\centering\arraybackslash}m{0.95cm}>{\centering\arraybackslash}m{0.95cm}>{\centering\arraybackslash}m{0.95cm}>{\centering\arraybackslash}m{0.95cm}>{\centering\arraybackslash}m{0.95cm}>{\centering\arraybackslash}m{0.95cm}>{\centering\arraybackslash}m{0.95cm}@{\hspace{13pt}}
 >{\centering\arraybackslash}m{0.95cm}>{\centering\arraybackslash}m{0.95cm}>{\centering\arraybackslash}m{0.95cm}} 
        \toprule

        \multirow{2}{*}{} 
         & \multicolumn{3}{c}{\textbf{\acs{csID}}}
         & \multicolumn{8}{c}{\textbf{\acs{nOOD}}}
         & \multicolumn{3}{c}{\textbf{\acs{fOOD}}} \\
        
        & 1$_{\text{b}}$ & 1$_{\text{c}}$ & Avg 
        & 2 & 3 & 4 & 5 & 6$_{\text{a}}$ & 6$_{\text{b}}$ & 7 & Avg 
        & CCAgT & FNAC & Avg\\
    \midrule
\rowcolor{Gray_light} \textbf{\acs{AUPRIN}}\,\raisebox{0.3ex}{$\uparrow$} & & & & & & & & & & & & & &\\
\rowcolor{Feature_light} MDSEns~\cite{lee2018simple_MDS} & 95.64 & 99.28 & 97.46 & 98.85 & 98.83 & 98.86 & 13.30 & 97.90 & 99.55 & 13.16 & 74.35 & 100.00 & 99.99 & 100.00 \\
\rowcolor{Feature_light} Residual~\cite{wang2022vim_ViM_Residual} & 10.35 & 13.69 & 12.02 & 19.55 & 6.85 & 12.35 & 10.88 & 8.92 & 34.45 & 8.00 & 14.43 & 55.84 & 72.06 & 63.95 \\
\rowcolor{Feature_light} MDS~\cite{lee2018simple_MDS} & 10.46 & 13.03 & 11.74 & 19.06 & 7.30 & 12.41 & 11.35 & 8.78 & 32.40 & 7.32 & 14.09 & 51.93 & 67.59 & 59.76 \\
\rowcolor{Combined_light} ViM~\cite{wang2022vim_ViM_Residual} & 14.44 & 17.27 & 15.85 & 21.00 & 10.45 & 12.41 & 11.78 & 7.80 & 33.97 & 7.31 & 14.96 & 62.33 & 76.03 & 69.18 \\
\rowcolor{Feature_light} KNN~\cite{sun2022out_KNN} & 11.81 & 11.15 & 11.48 & 15.55 & 11.42 & 9.60 & 10.33 & 6.15 & 31.42 & 6.33 & 12.97 & 28.37 & 65.82 & 47.10 \\
\bottomrule
    \end{tabular}}\par}
\end{table*}

\begin{table*}[htbp]
    \caption{Results from the \acs{PHAKIR} benchmark for the \acs{AUROC} and FPR@95 metrics. 
    M. Smoke and H. Smoke stands for Medium and Heavy Smoke.
    CAT. stands for CATARACTS.}
    \label{tab:phakir_auroc_fpr_vim_nnguide_corrected}
\centering
{\renewcommand\baselinestretch{1.0}\selectfont\resizebox{\textwidth}{!}{
\begin{tabular}{l@{\hspace{13pt}}>{\centering\arraybackslash}m{1.51cm}>{\centering\arraybackslash}m{1.51cm}>{\centering\arraybackslash}m{1.51cm}@{\hspace{13pt}}>{\centering\arraybackslash}m{1.51cm}>{\centering\arraybackslash}m{1.51cm}>{\centering\arraybackslash}m{1.51cm}>{\centering\arraybackslash}m{1.51cm}@{\hspace{13pt}}>{\centering\arraybackslash}m{1.51cm}>{\centering\arraybackslash}m{1.51cm}>{\centering\arraybackslash}m{1.51cm}} 
        \toprule

        \multirow{2}{*}{} 
         & \multicolumn{3}{c}{\textbf{\acs{csID}}}
         & \multicolumn{4}{c}{\textbf{\acs{nOOD}}}
         & \multicolumn{3}{c}{\textbf{\acs{fOOD}}} \\
        
        & {M. Smoke} & {H. Smoke} & Avg 
        & \acs{cholec} & \acs{EndoSeg15} & \acs{EndoSeg18} & Avg 
        & Kvasir & CAT. & Avg\\
    \midrule
\rowcolor{Gray_light} \textbf{\acs{AUROC}}\,\raisebox{0.3ex}{$\uparrow$} & & & & & & & & & & \\
\rowcolor{Feature_light} MDSEns~\cite{lee2018simple_MDS} & 45.67 & 84.43 & 65.05 & 96.65 & 94.81 & 99.87 & 97.11 & 99.98 & 97.02 & 98.50 \\
\rowcolor{Feature_light} Residual~\cite{wang2022vim_ViM_Residual} & 39.11 & 75.13 & 57.12 & 59.53 & 75.98 & 95.46 & 76.99 & 48.33 & 66.30 & 57.31 \\
\rowcolor{Feature_light} MDS~\cite{lee2018simple_MDS} & 38.25 & 73.84 & 56.04 & 58.27 & 76.37 & 94.81 & 76.48 & 40.20 & 62.75 & 51.47 \\
\rowcolor{Combined_light} ViM~\cite{wang2022vim_ViM_Residual} & 23.57 & 66.16 & 44.86 & 65.48 & 66.99 & 55.63 & 62.70 & 32.96 & 53.54 & 43.25 \\
\rowcolor{Feature_light} KNN~\cite{sun2022out_KNN} & 24.01 & 44.09 & 34.05 & 64.17 & 61.22 & 40.92 & 55.44 & 31.22 & 44.28 & 37.75 \\
\bottomrule
    \end{tabular}}\par}
\end{table*}

\begin{table*}[htbp]
    \caption{Results from the \acs{PHAKIR} benchmark for the \acs{AUPRIN} and \acs{AUPROUT} metrics. M. Smoke and H. Smoke stands for Medium and Heavy Smoke.
    CAT. stands for CATARACTS.}
\centering
{\renewcommand\baselinestretch{1.0}\selectfont\resizebox{\textwidth}{!}{
\begin{tabular}{l@{\hspace{13pt}}>{\centering\arraybackslash}m{1.51cm}>{\centering\arraybackslash}m{1.51cm}>{\centering\arraybackslash}m{1.51cm}@{\hspace{13pt}}>{\centering\arraybackslash}m{1.51cm}>{\centering\arraybackslash}m{1.51cm}>{\centering\arraybackslash}m{1.51cm}>{\centering\arraybackslash}m{1.51cm}@{\hspace{13pt}}>{\centering\arraybackslash}m{1.51cm}>{\centering\arraybackslash}m{1.51cm}>{\centering\arraybackslash}m{1.51cm}} 
        \toprule

        \multirow{2}{*}{} 
         & \multicolumn{3}{c}{\textbf{\acs{csID}}}
         & \multicolumn{4}{c}{\textbf{\acs{nOOD}}}
         & \multicolumn{3}{c}{\textbf{\acs{fOOD}}} \\
        
        & {M. Smoke} & {H. Smoke} & Avg 
        & \acs{cholec} & \acs{EndoSeg15} & \acs{EndoSeg18} & Avg 
        & Kvasir & CAT. & Avg\\
    \midrule
\rowcolor{Gray_light} \textbf{\acs{AUPRIN}}\,\raisebox{0.3ex}{$\uparrow$} & & & & & & & & & & \\
\rowcolor{Feature_light} MDSEns~\cite{lee2018simple_MDS} & 29.60 & 69.21 & 49.41 & 73.26 & 98.59 & 99.80 & 90.55 & 99.96 & 87.55 & 93.75 \\
\rowcolor{Feature_light} Residual~\cite{wang2022vim_ViM_Residual} & 27.14 & 52.52 & 39.83 & 1.16 & 92.18 & 94.58 & 62.64 & 32.41 & 1.11 & 16.76 \\
\rowcolor{Feature_light} MDS~\cite{lee2018simple_MDS} & 26.85 & 51.74 & 39.29 & 1.10 & 92.31 & 93.55 & 62.32 & 27.96 & 1.02 & 14.49 \\
\rowcolor{Combined_light} ViM~\cite{wang2022vim_ViM_Residual} & 22.71 & 40.13 & 31.42 & 1.26 & 86.48 & 38.06 & 41.93 & 25.88 & 0.38 & 13.13 \\
\rowcolor{Feature_light} KNN~\cite{sun2022out_KNN} & 22.77 & 27.39 & 25.08 & 1.41 & 84.54 & 32.37 & 39.44 & 26.53 & 0.33 & 13.43 \\
\bottomrule
    \end{tabular}}\par}
\end{table*}

\begin{table*}[htbp]
    \caption{Results from the \acs{OASIS3} benchmark for the \acs{AUROC} and FPR@95 metrics. }
\centering
{\renewcommand\baselinestretch{1.0}\selectfont\resizebox{\textwidth}{!}{
\begin{tabular}{l@{\hspace{13pt}}>{\centering\arraybackslash}m{1.51cm}>{\centering\arraybackslash}m{1.51cm}>{\centering\arraybackslash}m{1.51cm}@{\hspace{13pt}}>{\centering\arraybackslash}m{1.51cm}>{\centering\arraybackslash}m{1.51cm}>{\centering\arraybackslash}m{1.51cm}>{\centering\arraybackslash}m{1.51cm}@{\hspace{13pt}}>{\centering\arraybackslash}m{1.51cm}>{\centering\arraybackslash}m{1.51cm}>{\centering\arraybackslash}m{1.51cm}} 
        \toprule

        \multirow{2}{*}{} 
         & \multicolumn{3}{c}{\textbf{\acs{csID}}}
         & \multicolumn{4}{c}{\textbf{\acs{nOOD}}}
         & \multicolumn{3}{c}{\textbf{\acs{fOOD}}} \\
        
        & Modality & Scanner & Avg 
        & \acs{ATLAS} & \acs{BRATS} & CT & Avg 
        & \acs{MSD-H} & \acs{CHAOS} & Avg \\
    \midrule
\rowcolor{Gray_light} \textbf{\acs{AUROC}}\,\raisebox{0.3ex}{$\uparrow$} & & & & & & & & & & \\
\rowcolor{Feature_light} MDSEns~\cite{lee2018simple_MDS} & 100.00 & 100.00 & 100.00 & 98.47 & 99.98 & 99.94 & 99.46 & 100.00 & 100.00 & 100.00 \\
\rowcolor{Feature_light} Residual~\cite{wang2022vim_ViM_Residual} & 100.00 & 93.94 & 96.97 & 95.38 & 94.72 & 100.00 & 96.70 & 100.00 & 100.00 & 100.00 \\
\rowcolor{Feature_light} MDS~\cite{lee2018simple_MDS} & 100.00 & 92.62 & 96.31 & 94.65 & 93.80 & 100.00 & 96.15 & 100.00 & 100.00 & 100.00 \\
\rowcolor{Combined_light} ViM~\cite{wang2022vim_ViM_Residual} & 100.00 & 97.63 & 98.82 & 98.19 & 97.01 & 100.00 & 98.40 & 100.00 & 100.00 & 100.00 \\
\rowcolor{Feature_light} KNN~\cite{sun2022out_KNN} & 100.00 & 97.56 & 98.78 & 96.25 & 96.71 & 100.00 & 97.66 & 100.00 & 100.00 & 100.00 \\
\bottomrule
    \end{tabular}}\par}
\end{table*}

\begin{table*}[htbp]
    \caption{Results from the \acs{OASIS3} benchmark for the \acs{AUPRIN} and \acs{AUPROUT} metrics. }
    \label{tab:oasis_aupr_vim_nnguide_corrected}
\centering
{\renewcommand\baselinestretch{1.0}\selectfont\resizebox{\textwidth}{!}{
\begin{tabular}{l@{\hspace{13pt}}>{\centering\arraybackslash}m{1.51cm}>{\centering\arraybackslash}m{1.51cm}>{\centering\arraybackslash}m{1.51cm}@{\hspace{13pt}}>{\centering\arraybackslash}m{1.51cm}>{\centering\arraybackslash}m{1.51cm}>{\centering\arraybackslash}m{1.51cm}>{\centering\arraybackslash}m{1.51cm}@{\hspace{13pt}}>{\centering\arraybackslash}m{1.51cm}>{\centering\arraybackslash}m{1.51cm}>{\centering\arraybackslash}m{1.51cm}} 
        \toprule
        \multirow{2}{*}{} 
         & \multicolumn{3}{c}{\textbf{\acs{csID}}}
         & \multicolumn{4}{c}{\textbf{\acs{nOOD}}}
         & \multicolumn{3}{c}{\textbf{\acs{fOOD}}} \\
        & Modality & Scanner & Avg 
        & \acs{ATLAS} & \acs{BRATS} & CT & Avg 
        & \acs{MSD-H} & \acs{CHAOS} & Avg \\
    \midrule
\rowcolor{Gray_light} \textbf{\acs{AUPRIN}}\,\raisebox{0.3ex}{$\uparrow$} & & & & & & & & & & \\
\rowcolor{Feature_light} MDSEns~\cite{lee2018simple_MDS} & 100.00 & 100.00 & 100.00 & 85.77 & 99.89 & 99.22 & 94.96 & 100.00 & 100.00 & 100.00 \\
\rowcolor{Feature_light} Residual~\cite{wang2022vim_ViM_Residual} & 100.00 & 97.54 & 98.77 & 93.47 & 87.17 & 100.00 & 93.55 & 100.00 & 100.00 & 100.00 \\
\rowcolor{Feature_light} MDS~\cite{lee2018simple_MDS} & 99.99 & 96.98 & 98.49 & 92.51 & 85.57 & 100.00 & 92.69 & 100.00 & 100.00 & 100.00 \\
\rowcolor{Combined_light} ViM~\cite{wang2022vim_ViM_Residual} & 100.00 & 98.95 & 99.47 & 95.50 & 89.45 & 100.00 & 94.99 & 100.00 & 100.00 & 100.00 \\
\rowcolor{Feature_light} KNN~\cite{sun2022out_KNN} & 100.00 & 98.93 & 99.46 & 92.93 & 88.62 & 100.00 & 93.85 & 100.00 & 100.00 & 100.00 \\
\bottomrule
    \end{tabular}}\par}
\end{table*}

\begin{table*}[htbp]
    \caption{Results from the \acs{IN1k} benchmark for the \acs{AUROC} and FPR@95 metrics.}
    \label{tab:imagenet_auroc_fpr95_vim_nnguide_corrected}
\centering
{\renewcommand\baselinestretch{1.0}\selectfont\resizebox{\textwidth}{!}{
 \begin{tabular}{l@{\hspace{13pt}}>{\centering\arraybackslash}m{1.45cm}>{\centering\arraybackslash}m{1.45cm}>{\centering\arraybackslash}m{1.45cm}>{\centering\arraybackslash}m{1.45cm}@{\hspace{13pt}}
 >{\centering\arraybackslash}m{1.45cm}>{\centering\arraybackslash}m{1.45cm}>{\centering\arraybackslash}m{1.45cm}@{\hspace{13pt}}
 >{\centering\arraybackslash}m{1.45cm}>{\centering\arraybackslash}m{1.45cm}>{\centering\arraybackslash}m{1.45cm}>{\centering\arraybackslash}m{1.45cm}} 
        \toprule

        \multirow{2}{*}{} 
         & \multicolumn{4}{c}{\textbf{\acs{csID}}}
         & \multicolumn{3}{c}{\textbf{\acs{nOOD}}}
         & \multicolumn{4}{c}{\textbf{\acs{fOOD}}} \\
        
        & IN-V2 \cite{recht2019imagenet} & IN-C \,~\cite{hendrycks2018benchmarking} & IN-R \,~\cite{hendrycks2021many} & Avg 
        & SSB-hard \cite{vaze2022openset} & NINCO \cite{bitterwolf2023inorout} & Avg 
        & iNaturalist \cite{van2018inaturalist} & Textures \cite{cimpoi2014describing} & OpenImage-O \cite{wang2022vim_ViM_Residual} & Avg\\
    \midrule
\rowcolor{Gray_light} \textbf{\acs{AUROC}}\,\raisebox{0.3ex}{$\uparrow$} & & & & & & & & & & &\\
\rowcolor{Feature_light} MDSEns~\cite{lee2018simple_MDS} & 51.15 & 76.74 & 74.87 & 67.58 & 48.30 & 60.66 & 54.48 & 56.42 & 93.30 & 73.82 & 74.51 \\
\rowcolor{Feature_light} Residual~\cite{wang2022vim_ViM_Residual} & 49.83 & 67.78 & 65.52 & 61.04 & 42.14 & 54.59 & 48.37 & 52.13 & 87.81 & 61.02 & 66.99 \\
\rowcolor{Feature_light} MDS~\cite{lee2018simple_MDS} & 51.72 & 70.80 & 69.78 & 64.10 & 48.50 & 62.38 & 55.44 & 63.67 & 89.80 & 69.27 & 74.25 \\
\rowcolor{Combined_light} ViM~\cite{wang2022vim_ViM_Residual} & 57.34 & 83.77 & 87.95 & 76.35 & 65.53 & 78.63 & 72.08 & 89.56 & 97.97 & 90.50 & 92.68 \\
\rowcolor{Feature_light} KNN~\cite{sun2022out_KNN} & 56.44 & 83.94 & 87.64 & 76.01 & 62.57 & 79.64 & 71.10 & 86.41 & 97.09 & 87.04 & 90.18 \\
\bottomrule
    \end{tabular}}\par}
\end{table*}

\begin{table*}[htbp]
    \caption{Results from the \acs{IN1k} benchmark for the \acs{AUPRIN} and \acs{AUPROUT} metrics. }
    \label{tab:imagenet_auprin_auprout_vim_nnguide_corrected}
\centering
{\renewcommand\baselinestretch{1.0}\selectfont\resizebox{\textwidth}{!}{
 \begin{tabular}{l@{\hspace{13pt}}>{\centering\arraybackslash}m{1.45cm}>{\centering\arraybackslash}m{1.45cm}>{\centering\arraybackslash}m{1.45cm}>{\centering\arraybackslash}m{1.45cm}@{\hspace{13pt}}
 >{\centering\arraybackslash}m{1.45cm}>{\centering\arraybackslash}m{1.45cm}>{\centering\arraybackslash}m{1.45cm}@{\hspace{13pt}}
 >{\centering\arraybackslash}m{1.45cm}>{\centering\arraybackslash}m{1.45cm}>{\centering\arraybackslash}m{1.45cm}>{\centering\arraybackslash}m{1.45cm}} 
        \toprule

        \multirow{2}{*}{} 
         & \multicolumn{4}{c}{\textbf{\acs{csID}}}
         & \multicolumn{3}{c}{\textbf{\acs{nOOD}}}
         & \multicolumn{4}{c}{\textbf{\acs{fOOD}}} \\
        
        & IN-V2 \cite{recht2019imagenet} & IN-C \,~\cite{hendrycks2018benchmarking} & IN-R \,~\cite{hendrycks2021many} & Avg 
        & SSB-hard \cite{vaze2022openset} & NINCO \cite{bitterwolf2023inorout} & Avg 
        & iNaturalist \cite{van2018inaturalist} & Textures \cite{cimpoi2014describing} & OpenImage-O \cite{wang2022vim_ViM_Residual} & Avg\\
    \midrule
\rowcolor{Gray_light} \textbf{\acs{AUPRIN}}\,\raisebox{0.3ex}{$\uparrow$} & & & & & & & & & & &\\
\rowcolor{Feature_light} MDSEns~\cite{lee2018simple_MDS} & 82.96 & 91.56 & 82.45 & 85.66 & 48.23 & 92.32 & 70.28 & 87.13 & 98.47 & 88.79 & 91.46 \\
\rowcolor{Feature_light} Residual~\cite{wang2022vim_ViM_Residual} & 82.20 & 90.00 & 75.91 & 82.70 & 42.63 & 90.66 & 66.65 & 84.35 & 98.17 & 81.52 & 88.01 \\
\rowcolor{Feature_light} MDS~\cite{lee2018simple_MDS} & 83.11 & 91.37 & 79.81 & 84.76 & 49.37 & 93.24 & 71.30 & 89.91 & 98.60 & 87.35 & 91.95 \\
\rowcolor{Combined_light} ViM~\cite{wang2022vim_ViM_Residual} & 85.14 & 95.03 & 91.50 & 90.56 & 65.67 & 96.49 & 81.08 & 97.59 & 99.74 & 96.58 & 97.97 \\
\rowcolor{Feature_light} KNN~\cite{sun2022out_KNN} & 84.66 & 95.01 & 90.80 & 90.16 & 62.33 & 96.74 & 79.53 & 96.66 & 99.61 & 95.04 & 97.10 \\
\bottomrule
    \end{tabular}}\par}
\end{table*}

\begin{table*}[htbp]
\caption{Overview of the automatically selected hyperparameters for the three \acp{MIB} and \ac{IN1k}.}
    \label{tab:hyperparameters_vim_nnguide_corrected}
\centering
{\renewcommand\baselinestretch{1.0}\selectfont\resizebox{\textwidth}{!}{
\begin{tabular}{lll@{\hspace{20pt}}ll@{\hspace{20pt}}ll@{\hspace{20pt}}ll}
\toprule
\multirow{2}{*}{} & \multicolumn{2}{c}{\ac{MIDOG}} & \multicolumn{2}{c}{\ac{PHAKIR}} & \multicolumn{2}{c}{\ac{OASIS3}} & \multicolumn{2}{c}{\ac{IN1k}} \\
  Method                & Parameter 1 & Parameter 2 & Parameter 1 & Parameter 2 & Parameter 1 & Parameter 2 & Parameter 1 & Parameter 2 \\ 
\midrule
MDSEns           &  0.0                  & --                 & 0.0014                 & --                 & 0.0 & -- & 0.0 & --          \\
Residual         &  128                   & --                 & 128                 & --                 & 64 & -- & 1024 & --               \\
ViM              &  128                  & --                 & 128                 & --                 & 256 & -- & 256 & --                \\
KNN              &  5                 & --                 & 50                 & --                 & 5 & -- & 200 & --                \\
SHE              &  cosine                 & --                  & cosine & --                  & cosine & -- & cosine & --                \\ 
Relation         &  8               & --                & 8                 & --                 & 8 & -- & 1 & --                \\
\bottomrule
    \end{tabular}}\par}
\end{table*}

\end{document}